\documentclass[]{fairmeta}
\microtypesetup{expansion=false}

\usepackage[utf8]{inputenc}
\usepackage[T1]{fontenc}
\usepackage{amsmath,amssymb}
\usepackage{booktabs}
\usepackage{graphicx}
\usepackage{float}
\usepackage{caption}
\usepackage{xcolor,url}
\definecolor{url_color}{RGB}{113,187,179}
\definecolor{cite_color}{RGB}{99,177,137}
\definecolor{best_color}{RGB}{26,109,103}
\newcommand{\best}[1]{\textbf{\textcolor{best_color}{#1}}}
\newcommand{\figref}[1]{\hyperref[#1]{Figure~\ref*{#1}}}
\newcommand{\figrefs}[2]{\hyperref[#1]{Figures~\ref*{#1}}--\hyperref[#2]{\ref*{#2}}}
\newcommand{\tabref}[1]{\hyperref[#1]{Table~\ref*{#1}}}
\newcommand{\secref}[1]{\hyperref[#1]{Sec.~\ref*{#1}}}
\newcommand{\secrefs}[2]{\hyperref[#1]{Sec.~\ref*{#1}}--\hyperref[#2]{\ref*{#2}}}
\newcommand{\Secref}[1]{\hyperref[#1]{Section~\ref*{#1}}}
\newcommand{\Secrefs}[2]{\hyperref[#1]{Sections~\ref*{#1}}--\hyperref[#2]{\ref*{#2}}}
\newcommand{\eqrefn}[1]{\hyperref[#1]{(\ref*{#1})}}
\newsavebox{\sidebox}
\newdimen\colh
\newcommand{\sideboxheight}{\dimexpr\ht\sidebox+\dp\sidebox\relax}
\hypersetup{colorlinks=true,linkcolor=cite_color,citecolor=cite_color,urlcolor=url_color}
\title{ZYT-World: A Real-Time Controllable World Model for Closed-Loop Autonomous-Driving Simulation}
\author{ZYT AI Team}
\abstract{%
Generative world models offer controllable and repeatable closed-loop simulation for end-to-end and vision--language--action driving policies, but production deployment exposes three unresolved requirements: faithfully reproducing a mixed fisheye--pinhole rig at native resolutions; reconciling causal, per-timestep interaction with long-horizon stability and low latency; and preserving scene identity when a location is revisited along a different trajectory.

We present ZYT-World, a single architecture that natively generates four fisheye views with field of view (FoV) > 180$^\circ$ and three pinhole views. Projection-specific Pl\"ucker adapters encode camera geometry, ego-motion adaptive layer normalization (AdaLN) provides global motion control, and a lightweight pixel-aligned layout conditions traffic participants and signals through instance-level boxes, headings, colors, and directions. Heterogeneous training combines full-rig geometric coverage with high-resolution detail. Teacher forcing (TF), causal consistency distillation (CD), self-rollout distribution matching distillation (DMD), and RigCritic transform a 40-step bidirectional teacher into a one-step, per-latent streaming generator, with RigCritic evaluating the full seven-view rig jointly. A 19M-parameter variational autoencoder decoder (TinyVAE), W8A8 quantization, and our inference engine reduce decoding, backbone, and incremental-execution costs, respectively. Finally, cross-trajectory pairs derived from real captures train a plug-in implicit-memory module that preserves place-specific evidence when a location is revisited.

On the internal multi-view test set, the one-step model retains more than \best{90\%} of the teacher's PSNR and SSIM, while FID, FVD, and LPIPS stay within \best{11\%} of the teacher. Under the generator-only timing in \figref{fig:2}, it is \best{107.7$\times$} faster than the 40-step bidirectional teacher. TinyVAE decodes \best{59.8$\times$} faster than the Wan decoder. 30 s rollouts and cross-trajectory revisits show the intended long-horizon and memory behavior. Videos and additional results are available at \href{https://zyt-aim.github.io/ZYT-World/}{zyt-aim.github.io/ZYT-World}.%
}

\begin{document}
\maketitle
\section{Introduction}\label{sec:1}

End-to-end and vision--language--action driving policies require controllable, repeatable, and affordable closed-loop evaluation. Road testing is expensive and sparsely samples safety-critical long-tail events, whereas open-loop metrics cannot capture compounding interaction errors. A generative world model instead synthesizes multi-view observations from visual history and planned ego-motion and alternates with the policy in the simulation loop~\citep{russell2025gaia2,nvidia2026omnidreams,zheng2026xworld,li2026fardrive}. ZYT-World conditions on per-timestep pose increments; raw control commands must first be mapped into this representation.

\subsection{Four requirements for a closed-loop world model}\label{sec:1.1}

Because generated observations directly feed the policy, closed-loop utility depends on four capabilities.

\textbf{Sensor interchangeability.} Camera count, projection, calibration, native resolution, and pixel statistics should match the production rig so domain gaps do not enter the policy as perception errors.

\textbf{Motion response and scene control.} Each step depends only on history and current ego-motion, while roads, agents, and traffic lights remain editable for counterfactual long-tail scenarios.

\textbf{Spatiotemporal consistency.} Geometry and layout must remain coherent across views, long rollouts, and revisits along different trajectories.

\begin{figure}[!ht]
\centering
\includegraphics[width=0.92\textwidth,height=0.70\textheight,keepaspectratio]{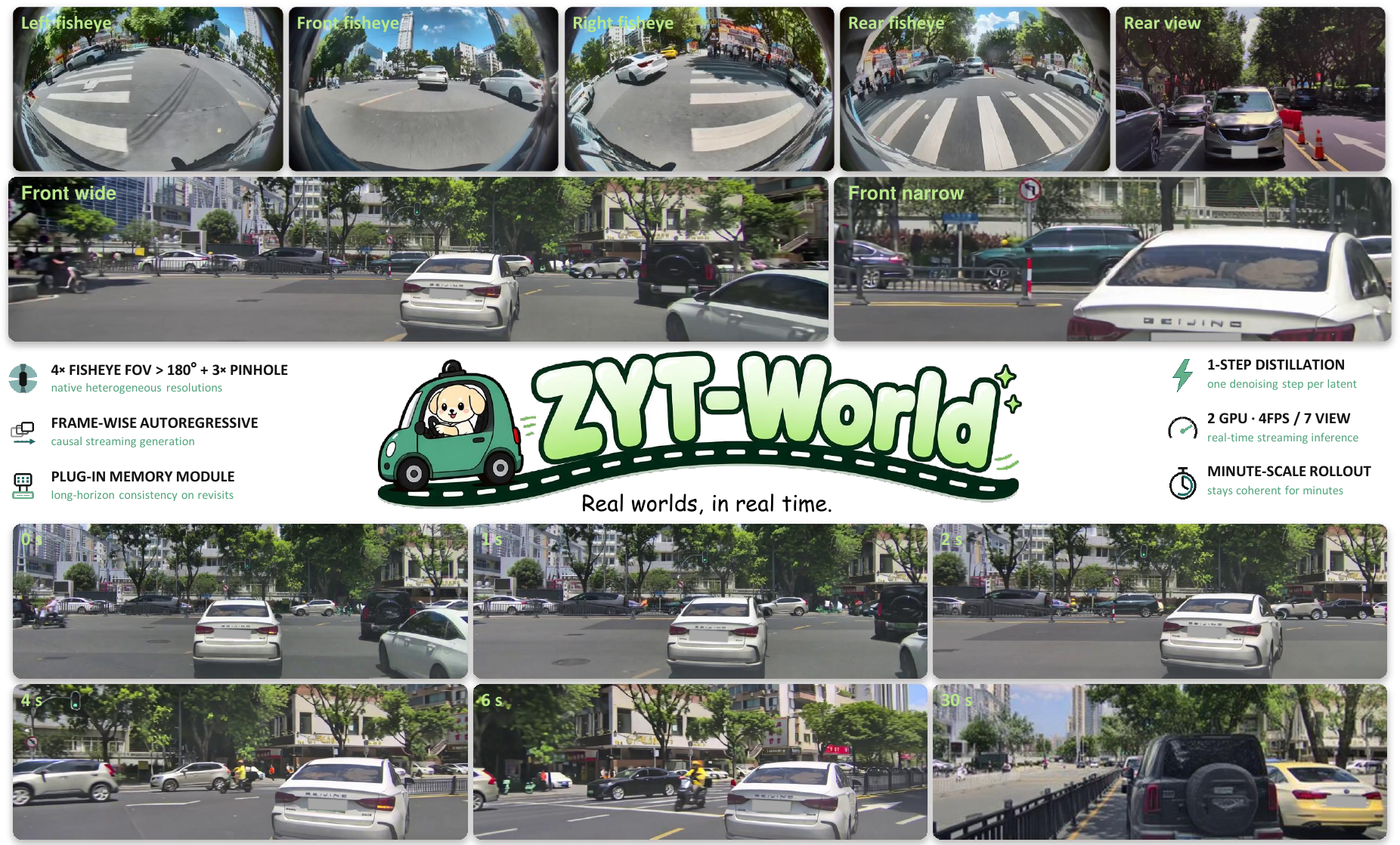}
\vspace{-1mm}
\caption{Native seven-view mixed fisheye--pinhole generation under per-latent, one-step, bounded-history rollout: 4 FPS for the full seven-view rig on two GPUs.}\label{fig:1}
\end{figure}

\textbf{Real-time deployability.} Generation plus decoding must fit the policy control period under a fixed compute budget. We report generator and decoder speedups separately; real-time throughput is 4 FPS for the full seven-view rig on the two-GPU setup in \figref{fig:1}.

\begin{figure}[!ht]
\centering
\includegraphics[width=0.98\textwidth,height=0.52\textheight,keepaspectratio]{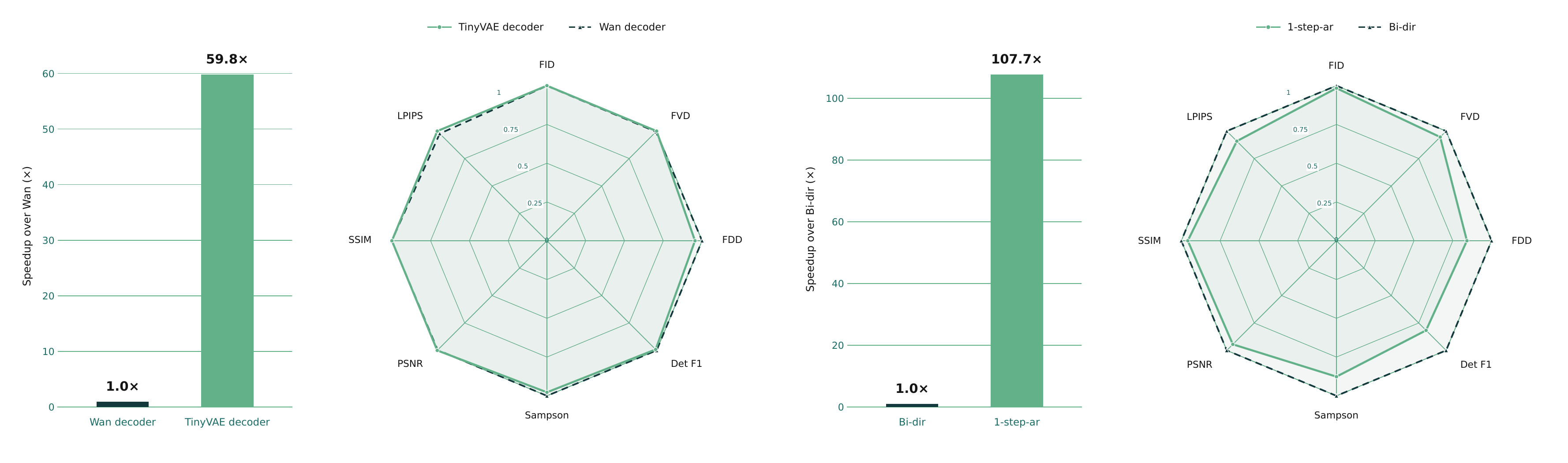}
\vspace{-1mm}
\caption{Decoder and generator efficiency: TinyVAE decoder versus Wan decoder (left), and one-step AR versus the 40-step bidirectional teacher (right). Per-metric trade-offs are in \secref{sec:7.2}.}\label{fig:2}
\end{figure}

Prior driving world models address autoregressive prediction, controllable multi-view diffusion, and long-horizon extrapolation~\citep{wen2024panacea,hu2023gaia1,wang2024drivedreamer,li2024drivingdiffusion,wang2024drivewm,lu2024wovogen,yang2024genad,zhao2025drivedreamer2,guo2024infinitydrive}. CausVid, Self Forcing, Causal Forcing++, ASD, One-Forcing, and AAD-1 further reduce the sampling cost of successive autoregressive units to a few steps or one step~\citep{yin2025causvid,huang2025selfforcing,zhao2026causalforcingpp,li2026aad1,yang2025asd,feng2026oneforcing}. ZYT-World targets the production setting in which four capabilities must coexist: native heterogeneous sensing, per-timestep control, cross-trajectory memory, and one-step streaming generation. \tabref{tab:1} places it against public driving world models along these axes.

\subsection{Three coupled problems under a production sensor rig}\label{sec:1.2}

Our production rig combines four cylindrical fisheye cameras with FoV > 180$^\circ$ and three pinhole cameras. All operate at approximately 720p, but their aspect ratios range from 5:1 to 5:4. This heterogeneity sharpens the preceding requirements into three coupled technical problems.

\textbf{Native heterogeneous sensing.} Fisheye and pinhole cameras have different ray geometry, and native resolutions yield unequal token sequences. Forcing a shared projection and resolution breaks interchangeability; modeling views independently weakens cross-view consistency. \secref{sec:3} treats geometry and control with projection-aware Pl\"ucker conditions, joint attention, and a unified dense interface; \secref{sec:5.1} treats mixed-resolution training.

\textbf{From bidirectional multi-step to causal one-step generation.} High-quality diffusion uses bidirectional attention and tens of evaluations, whereas a closed loop requires causal per-timestep response and low latency; step reduction also magnifies rollout error. \secref{sec:4} combines TF, CD, DMD, and RigCritic; \secrefs{sec:5}{sec:6} control training and inference cost through distributed execution, TinyVAE, W8A8, and our inference engine.

\textbf{Generation versus reconstruction.} A generative prior redraws a revisited road, and real logs rarely pair two trajectories through the same place. \secref{sec:2.4} renders novel trajectories from real-scene 4D Gaussian Splatting (4DGS), and \secref{sec:3.6} adds a zero-initialized plug-in memory path that restores place-specific geometry and is inherited by the autoregressive (AR) student through distillation.

\begin{table}[t]
\centering
\caption{Public driving world models versus ZYT-World on camera rig, resolution, interaction granularity, long-horizon rollout, and few-step generation.}\label{tab:1}
\scriptsize
\setlength{\tabcolsep}{3pt}
\renewcommand{\arraystretch}{1.08}
\resizebox{\textwidth}{!}{%
\begin{tabular}{l c c c c c }
\toprule
Work & Camera rig & Resolution & Interaction granularity & Long horizon (minute-scale) & Few-step generation \\
\midrule
Vista~\citep{gao2024vista} & Front only & 576$\times$1024 & Full clip & $\times$ & $\times$ \\
MagicDrive-V2~\citep{gao2025magicdrivev2} & 6 pinhole & Up to 848$\times$1600 & Full clip & $\times$ & $\times$ \\
GAIA-2 (Wayve)~\citep{russell2025gaia2} & 5 cameras & Unified 448$\times$960 & Full clip & $\times$ & $\times$ \\
Epona~\citep{zhang2025epona} & Front only & 512$\times$1024 & Frame-wise & $\checkmark$ & $\times$ \\
X-World~\citep{zheng2026xworld} & 7 cameras (1 fisheye) & Unreported & Chunk-wise & $\times$ & 4 steps \\
FAR-Drive~\citep{li2026fardrive} & Pinhole (nuScenes) & Unified 576$\times$1024 & Frame-wise & $\times$ & $\times$ \\
HorizonDrive~\citep{zhang2026horizondrive} & Pinhole (nuScenes) & Unified 256$\times$512 / 384$\times$768 & Chunk-wise & $\checkmark$ & 4 steps \\
OmniDreams (NVIDIA)~\citep{nvidia2026omnidreams} & 4 cameras, no fisheye & Unified 704$\times$1280 & Chunk-wise & $\checkmark$ & 2 steps \\
\best{ZYT-World (ours)} & \best{4 fisheye (FoV>180$^\circ$) + 3 pinhole} & \best{\textasciitilde{}720p, native 5:1--5:4} & \best{Frame-wise}\footnotemark & \best{$\checkmark$} & \best{1 step} \\
\bottomrule
\end{tabular}%
}
\end{table}
\footnotetext{Frame-wise (one VAE latent timestep) and chunk-wise (several) follow Causal Forcing++~\citep{zhao2026causalforcingpp}.}

ZYT-World further extends the deployment stack to decoding and scene reconstruction. TinyVAE combines pixel, perceptual, and semantic supervision to approach the decoding quality of Wan at a fraction of the cost, while the memory path restores place-specific geometry on revisits. We report generator-only and end-to-end pipeline latency separately.

Our main contributions toward controllable, long-horizon, and place-consistent generation on a production heterogeneous rig are:

\begin{itemize}
\item \textbf{Native heterogeneous-rig modeling (\secref{sec:3}, \secref{sec:5.1}).} Dual Pl\"ucker adapters, ego-motion AdaLN, and pixel-aligned layout jointly model four fisheye and three pinhole views without homogenizing projection or resolution.
\item \textbf{Causal one-step post-training (\secref{sec:4}).} TF, CD, DMD, and RigCritic convert a 40-step bidirectional teacher into a per-latent streaming generator whose bounded-history loop extends to minute-scale inference.
\item \textbf{Plug-in implicit memory (\secref{sec:2.4}, \secref{sec:3.6}).} Real-scene 4DGS provides cross-trajectory pairs; a zero-initialized residual path restores place-specific geometry and transfers to the AR student.
\item \textbf{Perceptually faithful lightweight decoding (\secref{sec:6.1}).} The 19M TinyVAE is trained with pixel, perceptual, and semantic supervision, yielding \best{59.8$\times$} faster decoding and \best{26.8$\times$} lower model memory with generation quality on par with the 555M Wan decoder.
\item \textbf{Training--inference co-design (\secref{sec:5}, \secrefs{sec:6.2}{sec:6.4}).} Multi-dimensional parallelism and memory optimization take seven-view HD training from out-of-memory to trainable on four GPUs; W8A8 and our inference engine cut per-forward and incremental-execution cost.
\end{itemize}

\begin{figure}[!ht]
\centering
\includegraphics[width=0.96\textwidth,height=0.62\textheight,keepaspectratio]{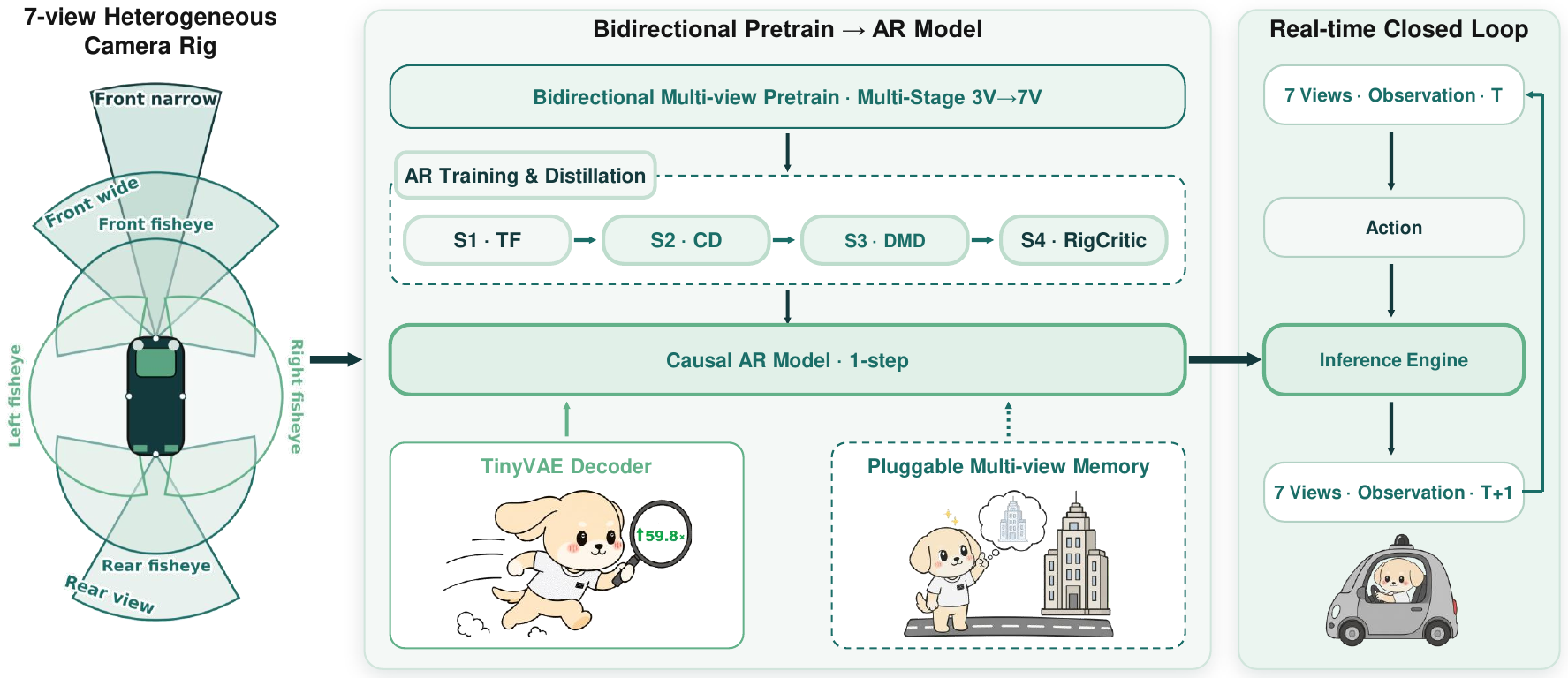}
\vspace{-1mm}
\caption{ZYT-World pipeline: from heterogeneous multi-view data and bidirectional pretraining, through autoregressive distillation, to real-time closed-loop inference.}\label{fig:3}
\end{figure}

The end-to-end system is summarized in \figref{fig:3}.

\section{Data system}\label{sec:2}

ZYT-World draws from the same onboard logs and capture pipeline used by the production driving stack, spanning multiple vehicle platforms. Every clip is collected under the compliance process of that stack and de-identified before use, and all subsequent processing and training run inside a controlled-access environment. This shared source aligns camera geometry, photometry, and perception-derived conditions with deployment. We do not preserve the raw fleet histogram: \secref{sec:2.3} deliberately rebalances weather, time of day, speed, and navigation actions. The following sections describe the rig and data scale (\secref{sec:2.1}), conditioning signals (\secref{sec:2.2}), cleaning and rebalancing (\secref{sec:2.3}), and cross-trajectory pairs for memory training (\secref{sec:2.4}).

\subsection{Sensor configuration and scale}\label{sec:2.1}

Each sample is an 81-frame, 10 FPS clip from seven synchronized surround cameras: front wide, front narrow, and rear view (pinhole), and front fisheye, left fisheye, right fisheye, and rear fisheye (cylindrical, FoV $>180^\circ$). The four fisheye cameras cover the near field; front wide, front narrow, and rear view cover the forward, far, and rear ranges. Every view retains the production rig's native resolution and aspect ratio (approximately 720p, from 5:1 to 5:4). We neither crop nor resample views to a common projection or resolution. The corpus contains millions of clips at two resolution tiers, supporting the staged curriculum in \secref{sec:5.2}.

\subsection{Conditioning signals}\label{sec:2.2}

The model consumes four condition families, frame-aligned with the video, taken from onboard capture and online perception. There is no extra manual annotation, so supervision scales with the fleet.

\begin{itemize}
\item \textbf{Ego-motion.} Per-frame ego-pose increments describe the vehicle trajectory and serve as the closed-loop motion condition (\secref{sec:3.4}).
\item \textbf{Camera parameters.} Calibrated extrinsics and intrinsics are stored with each clip and used to build Pl\"ucker ray fields (\secref{sec:3.3}), enabling unified modeling across vehicle platforms and camera layouts.
\item \textbf{Scene layout.} Dynamic agents and static road elements come from the online outputs of onboard perception, not offline HD labels. Trucks, cars, motorcycles, bicycles, and pedestrians use a fixed color per class; each 3D box also encodes position, size, and heading, so agents of the same class remain directionally editable. Traffic lights are localized by layout boxes, with markers encoding color and direction, enabling per-frame control of right-of-way. Light state directly governs stop, go, and turn decisions and is a key output of traffic-light perception, so fine alignment matters for closed-loop evaluation. The condition scales linearly with capture, but online perception inevitably produces misses, false positives, and jitter, so the model must be robust to condition noise (\secref{sec:3.5}).
\item \textbf{Text.} Unlike general video generators that rely on dense scene captions, text here \textbf{carries no scene semantics}: it is only a camera identifier. Weather, lighting, and content are learned from pixels, avoiding large-scale VLM labeling.
\end{itemize}

\begin{figure}[ht]
\centering
\includegraphics[width=0.96\textwidth,height=0.42\textheight,keepaspectratio]{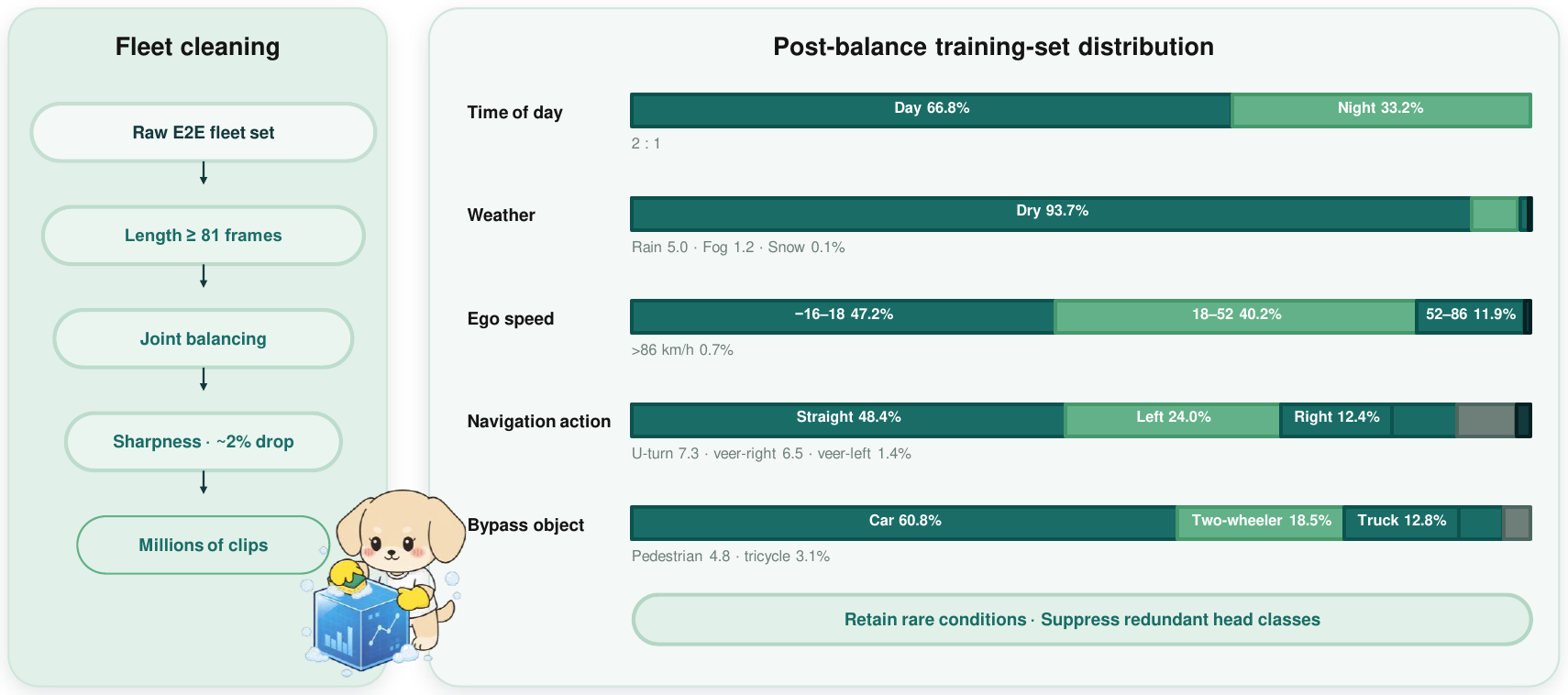}
\vspace{-1mm}
\caption{Data cleaning pipeline and the rebalanced training distribution.}\label{fig:4}
\end{figure}

\subsection{Cleaning and rebalancing}\label{sec:2.3}

Data preparation comprises duration filtering, distribution rebalancing, and sharpness filtering (\figref{fig:4}). We first remove clips shorter than 81 frames. Rebalancing provides the largest reduction because fleet logs are strongly imbalanced: dry weather dominates rain, fog, and snow, daytime dominates night, and cruising dominates turns and lane changes. We jointly stratify weather, time of day, ego speed, and navigation action, retaining rare conditions while subsampling redundant common cases. In the resulting corpus, the day-to-night ratio is approximately 2:1; dry weather (clear, cloudy, and overcast) accounts for 93.7\%, with rain, fog, and snow totaling about 6.3\%; speeds above 86 km/h contribute less than 1\%; straight driving accounts for nearly half of navigation actions, and left turns occur about twice as often as right turns; interacting agents are predominantly motor vehicles, with pedestrians and tricycles remaining rare. Finally, sharpness filtering removes approximately 2\% of clips severely degraded by lens contamination or heavy precipitation. These statistics characterize the rebalanced training distribution, not generation quality within every tail category.

\subsection{Training data for the memory module}\label{sec:2.4}

When the ego vehicle revisits a road segment along a different trajectory, static geometry and object placement should remain consistent with prior observations. A rolling KV-cache alone cannot provide this behavior: once a place leaves its finite history window, a later visit is redrawn from the generative prior. ZYT-World therefore adds an explicit memory condition for place-specific evidence (\secref{sec:3.6}). Training this path requires paired observations of the same location from different trajectories.

\begin{figure}[!ht]
\centering
\includegraphics[width=0.96\textwidth,height=0.62\textheight,keepaspectratio]{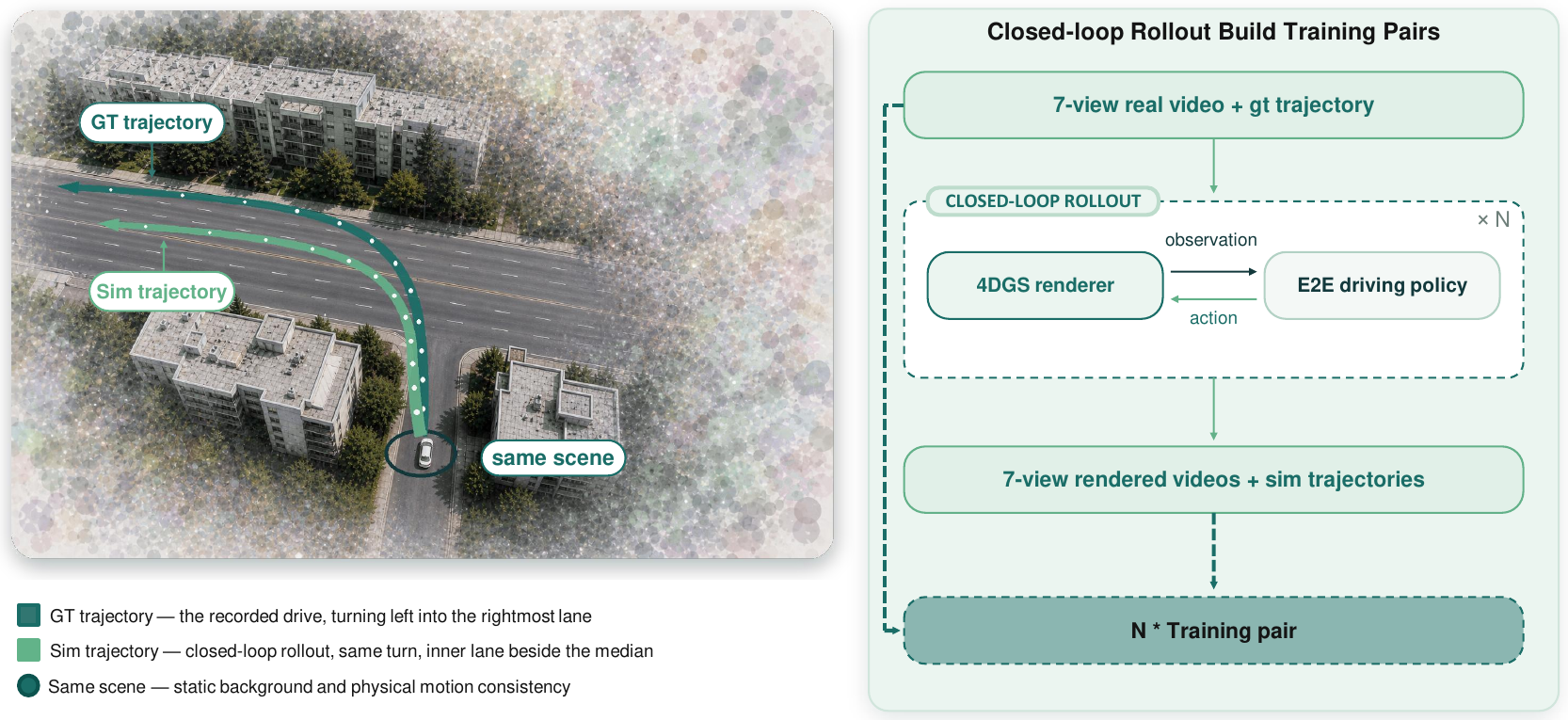}
\vspace{-1mm}
\caption{Construction of cross-trajectory memory training pairs.}\label{fig:5}
\end{figure}

Such supervision requires synchronized multi-view observations of the same scene from different trajectories, which production logs rarely provide as paired examples. We synthesize the missing view by having an end-to-end driving model propose a safety-constrained alternative path, reconstructing the captured dynamic scene with 4DGS~\citep{wu20244dgs}, and rendering that path. We use the formulation of \citet{wu20244dgs}, in which a spatiotemporal voxel grid and an MLP predict time-varying deformations of canonical 3D Gaussians~\citep{kerbl20233dgs}. The renderings form the memory input, while observations from the original onboard trajectory provide the target, yielding same-location, different-trajectory pairs. Both branches originate from one real capture and thus avoid game-engine appearance, although the rendered memory branch does not exactly match real-image memory at inference. We construct nearly one million such pairs (\figref{fig:5}).

\section{One backbone: native views, dense control, pluggable memory}\label{sec:3}

\begin{figure}[!ht]
\centering
\includegraphics[width=\textwidth,height=0.74\textheight,keepaspectratio]{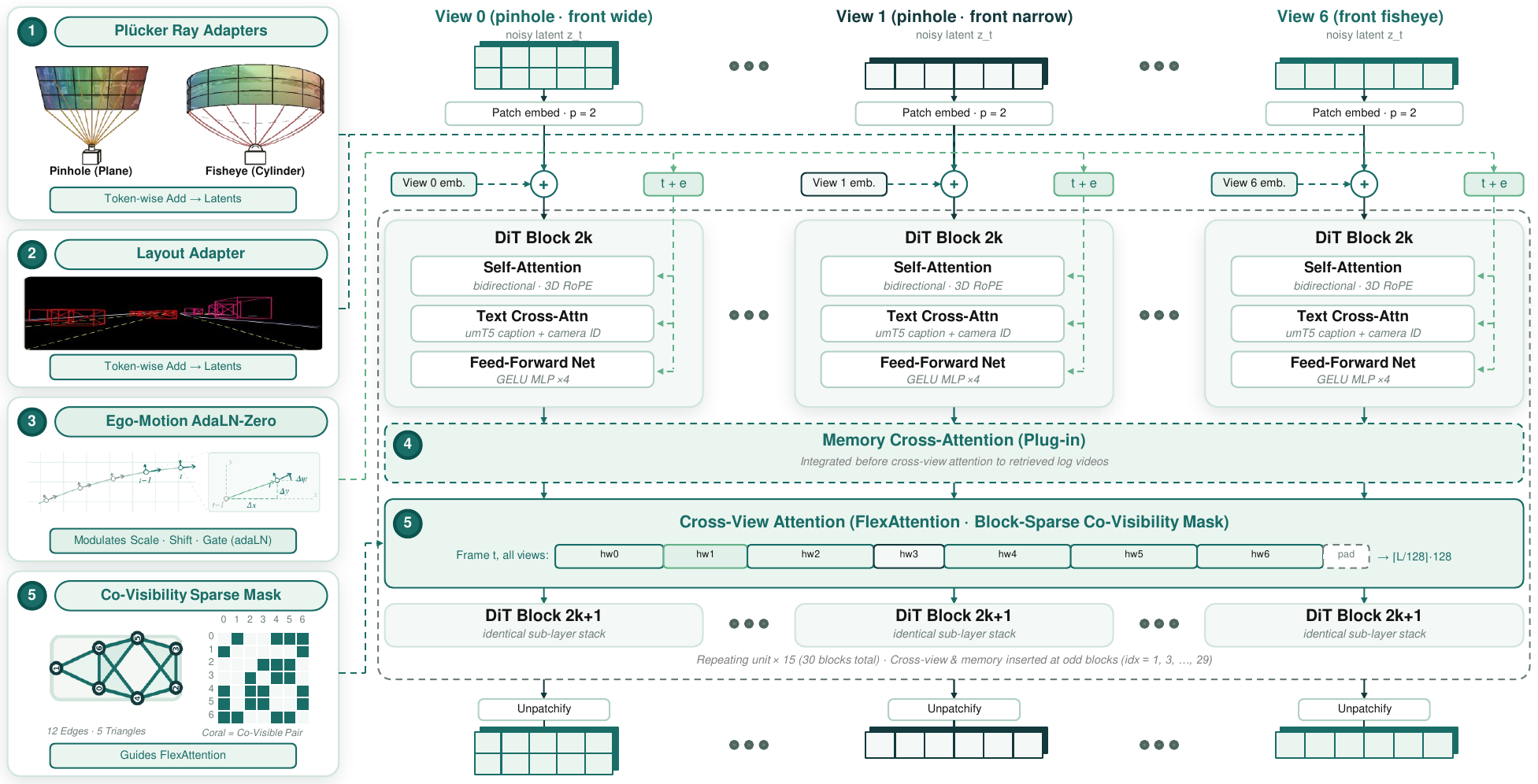}
\vspace{-1mm}
\caption{Heterogeneous multi-view controllable DiT with plug-in memory.}\label{fig:6}
\end{figure}

We parameterize ZYT-World as a single 5B-parameter video Diffusion Transformer (DiT)~\citep{peebles2023dit} that operates on a camera rig with heterogeneous projections and resolutions and accepts dense, pixel-aligned control (\figref{fig:6}). Within each view, spatiotemporal self-attention models temporal evolution; cross-view attention fuses spatial tokens at the same latent timestep (\secref{sec:3.1}). Camera rays and layout wireframes share a dense token-level injection interface (\secref{sec:3.2}), instantiated for camera geometry in \secref{sec:3.3} and scene layout in \secref{sec:3.5}. Ego-motion enters the timestep embedding and modulates every layer through AdaLN (\secref{sec:3.4}). A residual memory path provides cross-trajectory scene context (\secref{sec:3.6}). Across these components, conditions are rendered on native pixel grids, aligned with latent tokens, and attached through zero-initialized residual adapters. This initialization preserves the pretrained mapping while allowing each condition to be enabled independently or composed with the others.

\subsection{Backbone and multi-view attention}\label{sec:3.1}

Each camera is encoded and patchified independently in three dimensions, preserving its native $T\times h_v\times w_v$ token grid without cross-view padding or resampling. Intra-view positions use three-axis RoPE over $(t,h,w)$, following the video backbone; rotary positional embeddings themselves follow \citet{su2024roformer}.

\textbf{Per-view DiT.} All cameras share a video DiT with spatiotemporal self-attention, text cross-attention, FFNs, and AdaLN-Zero. Timestep embeddings are token-wise to support the per-frame noise schedules of \secref{sec:4.1}; AdaLN carries ego-motion, while text only identifies the camera.

\textbf{Semantic slots.} A seven-slot table $E\in\mathbb{R}^{7\times d}$, indexed by camera identity rather than batch position, supplies view identity and lets 3V/7V subset training share parameters.

\textbf{Geometry.} Fisheye and pinhole grid coordinates are not comparable, so cross-view attention omits RoPE; token-level Pl\"ucker rays (\secref{sec:3.3}) and view slots encode ray geometry and camera identity.

\textbf{Cross-view attention.} Fifteen independent modules are inserted every two layers and join spatial tokens only at the same latent timestep; temporal dependence remains in per-view attention. This reduces each sequence from $T\sum_v h_vw_v$ to $\sum_v h_vw_v$. Let $X_{v,t}\in\mathbb{R}^{h_vw_v\times d}$ be the tokens of camera $v$ at latent time $t$. Concatenation keeps native segment lengths:

\begin{equation}
Z_t=\big[\,X_{1,t}\,\|\,X_{2,t}\,\|\cdots\|\,X_{7,t}\,\big]\in\mathbb{R}^{L\times d},\qquad L=\sum_{v=1}^{7}h_v w_v.
\label{eq:cva-concat}
\end{equation}

Queries, keys and values $Q,K,V$ are linear maps of $Z_t$ with head dimension $d_h$. AdaLN supplies a residual scale $\gamma$; $W_o$ is the attention output projection and $W_0$ a zero-initialized residual projection. Token $i$ belongs to camera $\mathrm{view}(i)$; $\mathcal N(v)$ is the set of cameras physically overlapping $v$. With additive mask $\log\mathcal{M}$ and Iverson indicator $\mathbb{1}[\cdot]$,

\begin{align}
\mathrm{CVA}(Z_t)&=Z_t+\gamma\odot W_{0}\!\left(W_{o}\,\mathrm{softmax}\!\left(\frac{QK^{\top}}{\sqrt{d_h}}+\log\mathcal{M}\right)V\right),\label{eq:cva}\\
\mathcal{M}_{ij}&=\mathbb{1}\!\left[\mathrm{view}(j)\in\mathcal{N}(\mathrm{view}(i))\right].\label{eq:cva-mask}
\end{align}

$\mathcal N(v)$ retains 24 of the 42 directed inter-view pairs with physical overlap, removing 18 pair evaluations (approximately 43\% of directed pairs). Intra-view mixing remains in per-view attention and is not part of this mask. The zero initialization $W_0=0$ makes \eqrefn{eq:cva} an identity at the start of training; $Q$, $K$, $V$, $W_o$ and qk-norm are copied from self-attention only to initialize the internal mapping.

The covisibility graph uses FlexAttention \texttt{BlockMask}~\citep{dong2025flexattention}. The concatenated sequence alone is padded to a multiple of 128; masks are cached by token lengths and topology, and fully invalid blocks are skipped. Kernel and sequence-parallel details are in \secref{sec:5.3}.

\subsection{Unified injection of dense conditions}\label{sec:3.2}

Camera geometry and scene layout are both dense, pixel-aligned conditions. We process them through a common interface: render each condition on the target view's pixel grid, fold it onto the latent time axis with the VAE's causal 4$\times$ packing, map it to the DiT token grid with a lightweight adapter, and add it to the patch embedding. A zero-initialized output projection makes the inserted adapter initially approximate the pretrained backbone. This design supports initialization from Wan-series video-generation weights~\citep{wan2025wan} while introducing control with minimal perturbation to the pretrained mapping.

Both paths must land on the same token grid, which imposes a hard constraint. Let data-side downsample be $d$, PixelUnshuffle stride inside the adapter $s$, patch kernel $k$, VAE spatial compression $r=16$ and DiT patch $p=2$. Then

\begin{equation}
d \cdot s \cdot k = r \cdot p = 32.
\label{eq:grid-align}
\end{equation}

The camera and layout adapters satisfy \eqrefn{eq:grid-align} through different factorizations. The camera path uses $(d,s,k)=(4,4,2)$, moving part of the downsampling to data preparation and reducing condition pixels by 16$\times$ (\secref{sec:3.3}). The layout path uses $(1,16,2)$, preserving full-resolution rendering and applying lossless PixelUnshuffle within the adapter (\secref{sec:3.5}). Temporal packing exactly mirrors the VAE's nonuniform causal grouping: the first latent represents one RGB frame, so the condition path replicates that frame three times to fill the first four-frame group; subsequent frames are folded in groups of four along the channel axis (camera: 6$\rightarrow$24 channels; layout: 3$\rightarrow$12). This operation preserves temporal samples and lets the adapter mix them linearly rather than average them. Conditions are added after patch embedding and before the DiT blocks, only to the noisy stream; in the dual-stream formulation of \secref{sec:4.1}, clean history receives no direct condition embedding. Let $g(v)\in\{\mathrm{pinhole},\mathrm{fisheye}\}$ be the projection family of view $v$, $\mathbf{P}^{v}$ its Pl\"ucker ray field, $\mathbf{L}^{v}$ its layout wireframe, and $\mathbf{z}^{v}_{t}$ the noisy latent at latent time $t$. With projection-specific camera adapters $\mathcal{A}^{g(v)}_{\mathrm{cam}}$ and a shared layout adapter $\mathcal{A}_{\mathrm{lay}}$,

\begin{equation}
\mathbf{h}^{v}_{\mathrm{noisy}} = \mathrm{PatchEmbed}(\mathbf{z}^{v}_{t}) + \mathcal{A}^{g(v)}_{\mathrm{cam}}(\mathbf{P}^{v}) + \mathcal{A}_{\mathrm{lay}}(\mathbf{L}^{v}).
\label{eq:dense-inject}
\end{equation}

\subsection{Projection-aware Pl\"ucker ray fields}\label{sec:3.3}

Camera conditioning must describe the observation ray associated with every pixel. Global text or low-dimensional ego-motion cannot provide this correspondence, particularly when the rig mixes projections. We adopt the six-dimensional Pl\"ucker line representation of Light Field Networks~\citep{sitzmann2021lfn} and encode each pixel ray as a six-channel tensor. We extend that representation to the production rig in two ways: unprojection follows either the pinhole or cylindrical-fisheye camera model, and the resulting direction vector remains unnormalized. The representation dimension is shared across cameras; projection and calibration differences live in the ray values, which is our use of~\citep{sitzmann2021lfn}, not a claim made there.

Construction begins with the pose chain. Per-frame IMU poses and calibrated static extrinsics define each camera pose $T_i$, which we immediately convert to the adjacent increment $\Delta T_i=T_{i-1}^{-1}T_i$. Both conditioning paths encode these frame-to-frame increments rather than absolute poses, so each condition states how the current frame moves relative to the previous one, matching the causality of per-frame autoregression (\secref{sec:4}) and avoiding any coupling between scene identity and a global map frame. For Pl\"ucker encoding, directions are unprojected in the camera frame, rotated by the increment, and paired with the incremental camera center as the line origin $\mathbf o$, which keeps $\mathbf o\times\mathbf d$ nonzero and comparable across timesteps.

Given pose, \textbf{pinhole and fisheye differ only in unprojection}, mapping each pixel center to a camera-frame direction:

\begin{itemize}
\item Pinhole: $\tilde{\mathbf{d}} = \big(\tfrac{u+0.5-c_x}{f_x},\ \tfrac{v+0.5-c_y}{f_y},\ 1\big)$
\item Cylindrical fisheye (angle horizontally, linear height vertically): $\theta=\tfrac{u+0.5-c_x}{f_x},\ h=\tfrac{v+0.5-c_y}{f_y},\ \tilde{\mathbf{d}}=(\sin\theta,\ h,\ \cos\theta)$
\end{itemize}

Let $R_i$ and camera center $\mathbf o$ be the rotation and translation of the increment $\Delta T_i$. After unprojection, both camera types rotate the camera-frame direction and assemble the same unnormalized Pl\"ucker coordinates

\begin{equation}
\mathbf{d}=R_i\tilde{\mathbf{d}},\qquad\mathbf{p}=\big[\mathbf{o}\times\mathbf{d};\,\mathbf{d}\big]\in\mathbb{R}^{6}.
\label{eq:plucker}
\end{equation}

Unit-length rays would still encode field of view through the pixel-to-angle mapping; retaining magnitude instead preserves the pinhole constraint $\tilde d_z=1$ and cylindrical norm $\|\tilde{\mathbf d}\|=\sqrt{1+h^2}$ in the camera frame. For cylindrical unprojection $\tilde{\mathbf d}=(\sin\theta,h,\cos\theta)$, $\cos\theta<0$ when $|\theta|>\pi/2$, representing rays that point behind the camera when FoV exceeds 180$^\circ$.

A shared six-dimensional representation need not imply a shared encoder. In the camera frame, unnormalized pinhole directions satisfy $\tilde d_z\equiv1$, whereas cylindrical directions satisfy $\|\tilde{\mathbf d}\|=\sqrt{1+h^2}$. Although the first relation does not remain invariant after rotation into the shared frame, the two families retain different moment and norm statistics. We therefore use one adapter per projection family and dispatch through the semantic camera slot, making projection type explicit rather than requiring the backbone to infer it.

\subsection{Ego-motion: global AdaLN modulation}\label{sec:3.4}

Dense Pl\"ucker rays encode per-pixel geometry, but they do not provide a compact, frame-level motion signal. Moreover, the camera path uses data-side downsampling $d=4$ to keep seven-view HD training within memory. We therefore inject planar ego-motion separately through AdaLN. The two paths are complementary: AdaLN provides global motion modulation, while Pl\"ucker rays preserve projection-specific spatial geometry.

The AdaLN path takes three planar degrees of freedom from the same per-frame increment $\Delta T_i$, never from absolute pose: longitudinal translation $t_x$, lateral translation $t_y$ and yaw $\psi$; roll, pitch and vertical motion are not explicit conditions. The three quantities differ by 2--3 orders of magnitude; each is mapped into $[-1,1]$ by a scaled symlog, then packed by the VAE temporal scheme into 12 dimensions per latent timestep, and passed through Fourier features and an MLP to a motion embedding $\mathbf{e}_{\mathrm{ego}}$, which is added to the diffusion-timestep embedding to produce per-layer AdaLN coefficients. Per-feature affine modulation originates in FiLM~\citep{perez2018film}; our backbone uses DiT's AdaLN-Zero form~\citep{peebles2023dit}. Because ego-motion is frame-global, we use it for full-frame scale/shift and leave per-pixel projection differences to the Pl\"ucker path. FiLM and AdaLN-Zero do not themselves prescribe this split~\citep{peebles2023dit,perez2018film}.

\subsection{Layout: pixel-aligned lightweight wireframe injection}\label{sec:3.5}

Prior work represents structured scene control with instance tokens or scenario maps~\citep{russell2025gaia2,nvidia2026omnidreams,zheng2026xworld}. For a production heterogeneous rig, the additional requirement is to preserve projection-specific pixel alignment without expanding the backbone. Building on HorizonDrive's lightweight condition mapping~\citep{zhang2026horizondrive}, we rasterize class, 3D extent and heading, and traffic-signal color and direction into a wireframe on each native camera grid. A shared adapter maps the rasterized condition once to the token grid and injects it at a single backbone location, bypassing the video VAE. After rasterization, adapter cost is independent of backbone depth and the number of instances; editing a condition requires re-rendering the wireframe, and object identity is not modeled explicitly.

Upstream rendering uses camera calibration at native resolution. Color encodes agent class; boxes and markers encode extent and heading; traffic-light boxes and state markers place signal color and direction at the corresponding instance. Changing pose, size, heading, or state yields frame-wise counterfactual conditions. Wireframes share video preprocessing and are downsampled only by lossless PixelUnshuffle (\secref{sec:3.2}, $(d,s,k)=(1,16,2)$). Pure black maps exactly to zero, so missing, sparse, or locally dropped constraints need no extra mask. Pixel alignment is especially consequential for signals: small lamps and arrows directly govern right-of-way and stop--go behavior.

\sbox{\sidebox}{\begin{minipage}[t]{0.50\columnwidth}
\vspace{0pt}
\figref{fig:7} compares mapping the same wireframe with a token-grid adapter versus the video VAE encoder: 47.2M versus 149.6M parameters and 2.5 TMAC versus 366 TMAC. The adapter costs approximately $147\times$ less compute despite only a $3.2\times$ difference in the number of parameters, because it avoids 3D convolution at pixel resolution.
\end{minipage}}
\colh=\sideboxheight
\sbox0{\begin{minipage}[t]{0.47\columnwidth}
\vspace{0pt}
\centering
\includegraphics[width=\linewidth]{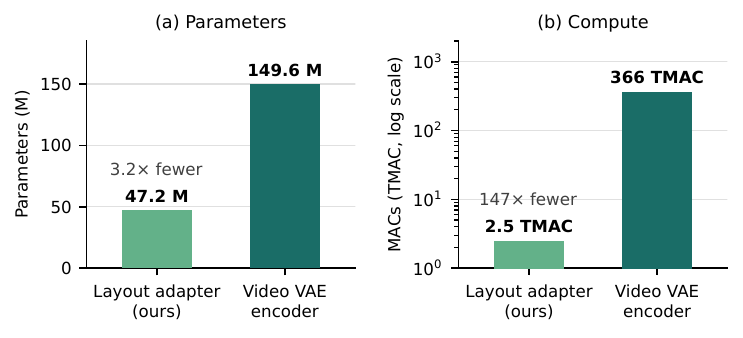}
\vspace{4pt}
{\small \figref{fig:7}: Compute cost of the layout path: lightweight adapter versus the video VAE encoder.\par}
\end{minipage}}
\ifdim\colh<\dimexpr\ht0+\dp0\relax \colh=\dimexpr\ht0+\dp0\relax \fi
\noindent
\begin{minipage}[t][\colh][s]{0.50\columnwidth}
\vspace{0pt}
\figref{fig:7} compares mapping the same wireframe with a token-grid adapter versus the video VAE encoder: 47.2M versus 149.6M parameters and 2.5 TMAC versus 366 TMAC. The adapter costs approximately $147\times$ less compute despite only a $3.2\times$ difference in the number of parameters, because it avoids 3D convolution at pixel resolution.

\vfill
Projection distortion is already rasterized into each wireframe, so all seven views can share an adapter composed of PixelUnshuffle, a linear projection, and a zero-initialized $1\times1$ convolution.
\end{minipage}\hfill
\begin{minipage}[t][\colh][s]{0.47\columnwidth}
\vspace{0pt}
\centering
\includegraphics[width=\linewidth]{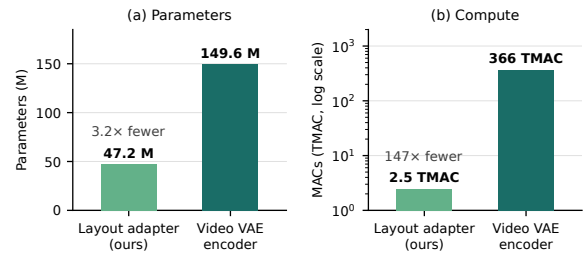}
\vfill
\captionof{figure}{Compute cost of the layout path: lightweight adapter versus the video VAE encoder.}\label{fig:7}
\end{minipage}
\par
Camera rays retain separate adapters because their input statistics differ by projection family. If wireframes are unavailable during training, we exclude the entire clip rather than mix conditioned and unconditioned views within one rig. At inference, missing timestamps use an all-zero layout to preserve tensor shape; this represents an unconstrained layout, not an inferred mask.

\subsection{Memory: long-horizon scene fidelity across trajectories}\label{sec:3.6}

\begin{figure}[!ht]
\centering
\includegraphics[width=0.96\textwidth,height=0.62\textheight,keepaspectratio]{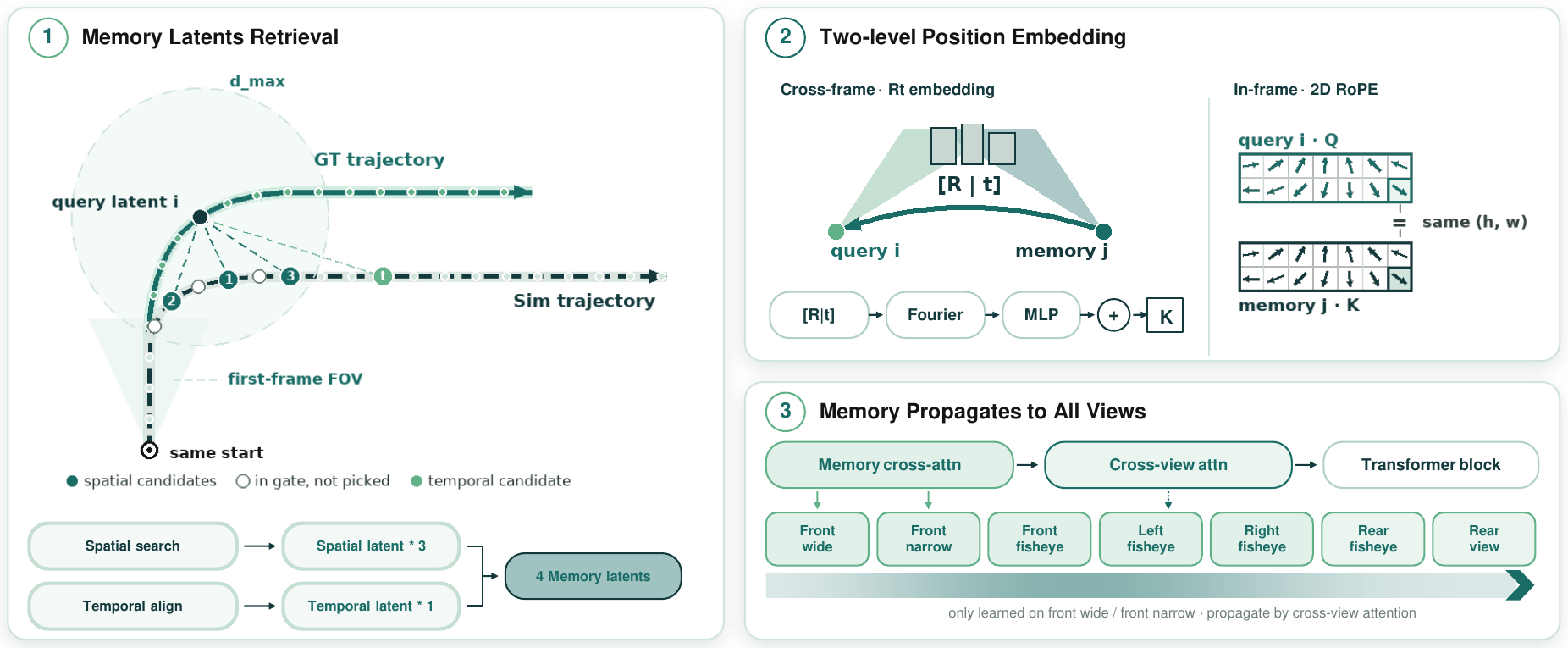}
\vspace{-1mm}
\caption{Memory-latent retrieval, two-level position embedding, and propagation to all views.}\label{fig:8}
\end{figure}

Autoregression extends time; memory preserves place-specific geometry and layout when the scene is revisited along another trajectory. The 4DGS pairs of \secref{sec:2.4} provide cross-trajectory supervision, while the plug-in implicit path avoids explicit reconstruction. \figref{fig:8} summarizes the pathway detailed below.

\textbf{Module.} Residual cross-attention is inserted before 15 odd backbone layers: target tokens query all tokens from four retrieved memory latents, batched per frame with FlashAttention~\citep{dao2022flashattention}. When no memory is retrieved the tokens are set to zeros, a mixed-data dropout that keeps generation well-behaved with or without memory, and the path is carried through the distillation of \secref{sec:4}.

\textbf{Memory Latents Retrieval.} For each target latent, 20 m and 45$^\circ$ gates filter candidates; heading then distance selects three spatial references separated by at least 1.5 m, and temporal alignment supplies the nearest-time reference. Selection is entirely data-side.

\textbf{Two-level Position Embedding.} Fourier--MLP features of relative rotation and translation encode cross-frame geometry, while 2D RoPE aligns pixels within each frame. Memory logs receive no absolute rollout time and always match the target patch grid.

\textbf{Memory Propagates to All Views.} Shared memory layers attach only to the front-wide and front-narrow slots and precede cross-view attention, which propagates retrieved evidence to the other five cameras without seven independent retrievals.

\section{From bidirectional to autoregressive}\label{sec:4}

This section turns a bidirectional, multi-step multi-view world model into a causal streaming generator for long-horizon generation and per-timestep interaction (the second problem in \secref{sec:1.2}). AR post-training proceeds in four phases, distinct from the pretraining curriculum of \secref{sec:5.2}: \secref{sec:4.1} TF aligns training with deployment-time causal visibility; \secref{sec:4.2} CD provides a few-step initialization; \secref{sec:4.3} DMD matches distributions on self-generated histories; and \secref{sec:4.5} stops DMD and applies RigCritic, an asymmetric adversarial refinement of the one-step generator. The pixel-domain perceptual loss in \secref{sec:4.4} is a generator regularizer used in DMD and RigCritic rather than a separate phase. All components share the same multi-view DiT, conditioning interface, and attention semantics. \Secref{sec:4.6} then describes bounded-memory inference for minute-scale streaming.

In VAE latent space, let $\mathbf{z}^{1:V}_{1:T}$ be the autoregressive context window of $V$ cameras and $\mathbf{c}$ the control (camera, ego-motion, and layout). The model predicts the next multi-view latent at a \textbf{latent timestep}---a frame-wise autoregressive unit in the sense of Causal Forcing++~\citep{zhao2026causalforcingpp}---and appends it to the history buffer:

\begin{equation}
\hat{\mathbf{z}}^{1:V}_{T+1} \sim p_\theta\big(\mathbf{z}^{1:V}_{T+1} \mid \mathbf{z}^{1:V}_{1:T},\ \mathbf{c}_{T+1}\big).
\label{eq:ar-step}
\end{equation}

\subsection{Causalization and TF}\label{sec:4.1}

At deployment, the KV-cache stores only denoised clean history; retaining noise on historical keys, as Diffusion Forcing permits~\citep{chen2024diffusionforcing}, would mismatch this interface. We concatenate two streams. Clean tokens $c_0,\ldots,c_{T-1}$ are ground-truth latents at diffusion timestep 0, carry no direct control features, and supply only keys and values. Noisy tokens $n_0,\ldots,n_{T-1}$ draw independent diffusion timesteps per latent and are the denoising targets. One or five RGB conditioning frames compress to $n_c\in\{1,2\}$ latents; the first $n_c$ noisy slots are replaced by ground truth, and the loss applies only for $f\ge n_c$. The autoregressive unit is one latent. Indices $i,j$ run over clean slots and $f,g$ over noisy slots. The attention mask is

\begin{equation}
\begin{aligned}
M(c_i,c_j)&=\mathbb{1}[j\le i],&
M(c_i,n_g)&=0,\\
M(n_f,n_g)&=\mathbb{1}[g=f],&
M(n_f,c_j)&=\mathbb{1}[j<f].
\end{aligned}
\label{eq:tf-mask}
\end{equation}

\begin{figure}[!ht]
\centering
\includegraphics[width=0.98\textwidth,height=0.52\textheight,keepaspectratio]{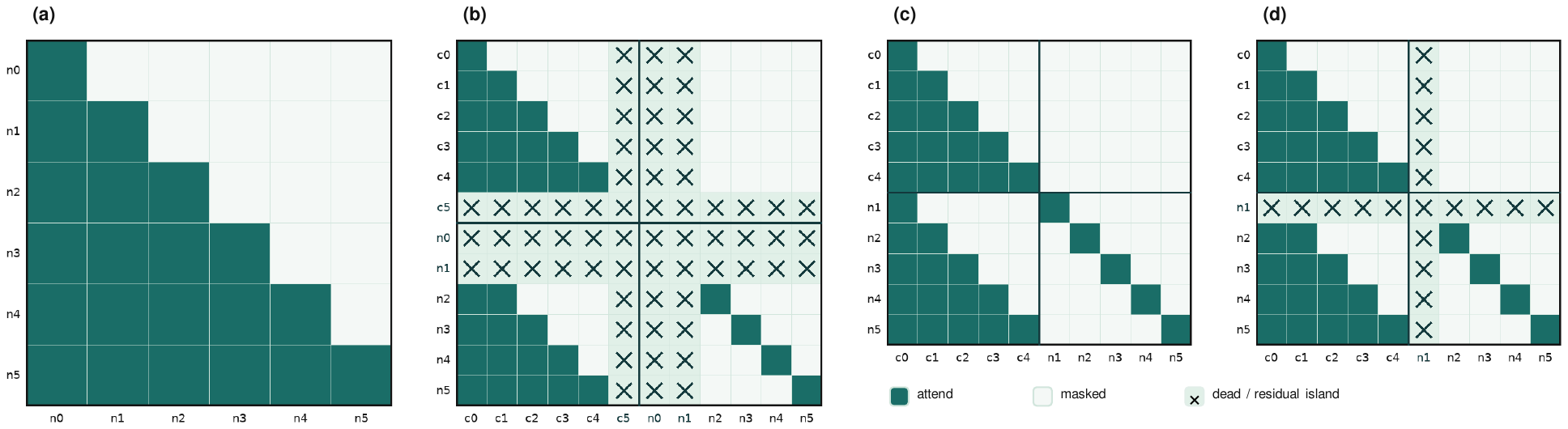}
\vspace{-1mm}
\caption{Teacher-forcing mask and condition-aware fixed-length pruning.}\label{fig:9}
\end{figure}

As shown in \figref{fig:9}, clean tokens attend only to causal clean history, including themselves; noisy tokens attend only to themselves and strictly earlier clean tokens. $c_{T-1}$ is never consumed as a key and is never supervised. The first $n_c$ noisy condition slots are also unsupervised: $n_1$ is unused when $n_c=2$ but is the first generated latent when $n_c=1$. Pruning every currently unused token would vary the sequence length and trigger recompilation, so we prune only the intersection of the unused sets, always $\{c_{T-1},n_0\}$, and retain $n_1$. Pruning precedes patchification and slices RoPE, timesteps, and all conditions identically, fixing the length at $2T-2$. For $T=21$, 42 latent entries become 40: 4.8\% fewer tokens and $\approx$0.91$\times$ quadratic-attention cost. Token-wise CPU fp64 and 12-step full-sequence checks match the unpruned path within $10^{-5}$.

Under rectified interpolation $\mathbf z_{\tau}=(1-\sigma_{\tau})\mathbf z_0+\sigma_{\tau}\boldsymbol\epsilon$ at per-latent noise levels $\{\tau_f\}$, training uses independent levels per latent and applies flow matching only to generated latents. Let $v_\theta$ be the model's velocity field and $w_v$ a per-view loss weight (uniform unless noted):

\begin{equation}
\mathcal{L}_{\mathrm{FM}}=\mathbb{E}\Big[\sum_v w_v\big\|v_\theta(\mathbf{z}^v_{\tau},\{\tau_f\},\mathbf{c})-(\boldsymbol{\epsilon}^v-\mathbf{z}^v_0)\big\|^2_{f\ge n_c}\Big].
\label{eq:fm}
\end{equation}

An optional bell-shaped timestep weight emphasizes intermediate noise levels, where conditioning is most informative. Pl\"ucker rays, ego-motion, and layout are injected per latent timestep from pretraining Stage 1 (front three-view), so controllability is learned jointly with generation rather than added after seven-view training.

\subsection{Causal CD}\label{sec:4.2}

Starting DMD from a multi-step causal model leaves a large optimization gap, while storing full probability-flow ODE trajectories is expensive. Following Causal Forcing++~\citep{zhao2026causalforcingpp}, we construct a few-step initialization with online CD. The shifted flow-matching grid is discretized into $N_{\mathrm{CD}}=48$ levels $t_0>t_1>\cdots>t_{N_{\mathrm{CD}}-1}$. Each iteration samples adjacent levels $(t_i,t_{i+1})$, and the frozen causal teacher flow $v_\phi$ takes one ODE step $\operatorname{Step}$, optionally with CFG. For AR unit $k$, $\bar{\mathbf c}_k=(\mathbf c,\mathbf z_{\mathrm{gt}}^{<k})$ combines the control with teacher-forced clean history:

\begin{equation}
\tilde{\mathbf z}^{k}_{t_{i+1}}=\operatorname{Step}\!\left(\mathbf z^{k}_{t_i},v_\phi(\mathbf z^{k}_{t_i},t_i,\bar{\mathbf c}_k),t_i\!\rightarrow\!t_{i+1}\right).
\label{eq:cd-step}
\end{equation}

Let $F_\theta(\mathbf z_t,t,\bar{\mathbf c}_k)$ denote the online student's clean-latent prediction obtained from its flow field, $\theta^-$ an exponential moving average of $\theta$, $\operatorname{sg}[\cdot]$ a stop-gradient, and $w(t_i)$ a positive timestep weight. Adjacent-level consistency is

\begin{equation}
\mathcal L_{\mathrm{CD}}=\mathbb E_i\!\left[w(t_i)\sum_v w_v\left\|F_\theta(\mathbf z^{v,k}_{t_i},t_i,\bar{\mathbf c}_k)-\operatorname{sg}\!\left[F_{\theta^-}(\tilde{\mathbf z}^{v,k}_{t_{i+1}},t_{i+1},\bar{\mathbf c}_k)\right]\right\|^2\right].
\label{eq:cd}
\end{equation}

Adjacent-level supervision decomposes full-trajectory regression into local consistency learning without pre-generating teacher trajectories. Frozen teacher, EMA student, and online student each process the full multi-view sequence in one forward and reuse the mask and fixed-length pruning of \secref{sec:4.1}, preserving covisible interaction and a static input shape. The result is a stable causal few-step initialization for self-rollout DMD.

\subsection{Self-rollout DMD for one-step generation}\label{sec:4.3}

Causal CD is trained on ground-truth history, whereas deployment conditions on the model's own predictions. We address this history shift by combining DMD/DMD2 distribution matching~\citep{yin2024dmd,yin2024dmd2}, the causal-student/bidirectional-teacher asymmetry of CausVid~\citep{yin2025causvid}, and the self-rollouts of Self Forcing~\citep{huang2025selfforcing}. AAPT pursues a similar train--inference alignment through student forcing and a training-time KV-cache in an adversarial framework~\citep{lin2025aapt}; here the cache is used within DMD. Starting from one clean conditioning latent, the student generates the remaining 20 latents in one step each, yielding $\hat{\mathbf z}_0=G_\theta(\boldsymbol\epsilon,\mathbf c)$ with an entirely self-generated history.

The generator $G_\theta$ is strictly causal, whereas the frozen teacher and the online fake score model both attend bidirectionally over the full seven-view clip. We re-noise the student rollout at a random diffusion level $\tau$ to obtain $\mathbf z_\tau$; $\mu_{\mathrm{real}}$ and $\mu_{\mathrm{fake}}$ denote the frozen teacher's and the online fake score model's clean-latent ($x_0$) predictions, not raw scores. The teacher may use classifier-free guidance with increment $\alpha_{\mathrm{cfg}}$, so the conventional guidance scale is $s=1+\alpha_{\mathrm{cfg}}$:

\begin{equation}
\mu_{\mathrm{real}}\leftarrow\mu^{\mathrm{cond}}_{\mathrm{real}}+\alpha_{\mathrm{cfg}}\!\left(\mu^{\mathrm{cond}}_{\mathrm{real}}-\mu^{\mathrm{uncond}}_{\mathrm{real}}\right).
\label{eq:cfg}
\end{equation}

With $g\propto\mu_{\mathrm{fake}}-\mu_{\mathrm{real}}$, the stop-gradient surrogate below moves $\hat{\mathbf z}_0$ toward the teacher prediction and away from the fake-model prediction, giving the usual $x_0$-parameterized proxy for reverse-KL distribution matching~\citep{yin2024dmd,yin2024dmd2}. After per-sample normalization over latent axes $(f,c,h,w)$ with stabilizer $\varepsilon>0$, the generator gradient is
\enlargethispage{3\baselineskip}
\nopagebreak
\begin{align}
\nabla_\theta\mathcal L_{\mathrm{DMD}}&=\mathbb E_{\tau,\boldsymbol\epsilon}\!\left[\left\langle g,\frac{\partial\hat{\mathbf z}_0}{\partial\theta}\right\rangle\right],\label{eq:dmd-grad}\\
g&=\frac{\mu_{\mathrm{fake}}(\mathbf z_\tau,\tau,\mathbf c)-\mu_{\mathrm{real}}(\mathbf z_\tau,\tau,\mathbf c)}{S},\label{eq:dmd-g}\\
S&=\operatorname{mean}_{f,c,h,w}\!\left(\left|\hat{\mathbf z}_0-\mu_{\mathrm{real}}(\mathbf z_\tau,\tau,\mathbf c)\right|\right)+\varepsilon.\label{eq:dmd-S}
\end{align}
\pagebreak

We implement the same update through an equivalent stop-gradient surrogate:

\begin{equation}
\mathcal L^{\mathrm{surr}}_{\mathrm{DMD}}=\frac{1}{2}\left\|\hat{\mathbf z}_0-\operatorname{sg}\!\left[\hat{\mathbf z}_0-g\right]\right\|_2^2.
\label{eq:dmd-surr}
\end{equation}

The generator and fake score model are updated alternately so that the latter tracks the evolving student distribution. On self-rollouts, the fake score $v_\psi$ uses the base flow-matching objective: diffusion level 0 for the clean prefix $\mathbf z_{\mathrm{cond}}$, $\tau$ for the generated region, and loss only on that region:

\begin{equation}
\mathcal L_{\mathrm{critic}}(\psi)=\mathbb E_{\tau}\!\left[\sum_v w_v\left\|\left[v_\psi\!\left([\mathbf z^v_{\mathrm{cond}},\mathbf z^v_\tau],\{0,\tau\},\mathbf c\right)\right]_{\mathrm{gen}}-\left(\boldsymbol\epsilon^v-\hat{\mathbf z}^v_0\right)\right\|_2^2\right].
\label{eq:critic}
\end{equation}

This DMD phase transfers clip-level distributional information from a bidirectional teacher to a strictly causal one-step generator while exposing the student to its own rollout history. It complements causal CD, which conditions on ground-truth prefixes, and provides the generator objective to which paired perceptual supervision is added.

\subsection{Pixel-domain perceptual regularization}\label{sec:4.4}

DMD aligns the global latent distribution, yet DMD-only long rollouts still show highlight bloom, blurred fine texture, and malformed small objects; the frozen teacher may itself drift from real observations. We therefore add a ground-truth-paired pixel-domain term to the generator: DMD constrains distribution, while perceptual supervision constrains local observation fidelity.

\begin{figure}[!ht]
\centering
\includegraphics[width=0.98\textwidth,height=0.52\textheight,keepaspectratio]{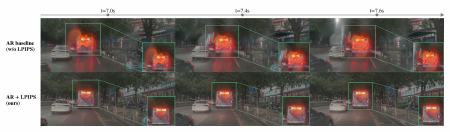}
\vspace{-1mm}
\caption{One-step AR generation without (top) and with (bottom) pixel-domain LPIPS regularization. Green boxes mark local differences.}\label{fig:10}
\end{figure}

Differentiable decoding of seven views at $T=21$ makes the original Wan2.2 VAE impractical inside the training loop. We instead freeze $D_{\mathrm{light}}$ from \secref{sec:6.1}, aligned to the Wan VAE's latent space and temporal upsampling. Prediction and target are prefixed by the same clean condition latent, decoded, and compared with frozen-AlexNet LPIPS~\citep{zhang2018lpips}. Let $\mathcal V_s$ be the sampled views, $\mathcal F$ the valid frames after decoder warmup, and $D_{\mathrm{light}}$ the frozen TinyVAE decoder of \secref{sec:6.1}:

\begin{equation}
\mathcal L_{\mathrm{perc}}=\frac{1}{|\mathcal F|\sum_{v\in\mathcal V_s}w_v}\sum_{v\in\mathcal V_s}w_v\sum_{f\in\mathcal F}\operatorname{LPIPS}\!\left(D_{\mathrm{light}}([\mathbf z^v_{\mathrm{cond}},\hat{\mathbf z}^v_{1:T-1}])_f,\,D_{\mathrm{light}}([\mathbf z^v_{\mathrm{cond}},\mathbf z^v_{1:T-1}])_f\right).
\label{eq:perc}
\end{equation}

The generator objective is

\begin{equation}
\mathcal L_G=\mathcal L^{\mathrm{surr}}_{\mathrm{DMD}}+\lambda_{\mathrm{perc}}\mathcal L_{\mathrm{perc}},\qquad\lambda_{\mathrm{perc}}=0.1.
\label{eq:gen-dmd}
\end{equation}

Both branches use the same conditioning latent as a decoder prefix, preventing decoder cold-start state from creating a spurious discrepancy. With decoder warm-up window $w_{\mathrm{dec}}=3$, the first $(w_{\mathrm{dec}}-1)\times4$ pixel frames are discarded. Decoding runs in groups of three latents with streamed loss accumulation; the decode--pixel--LPIPS chain is activation-checkpointed, making peak memory approximately independent of rollout length. Each iteration samples three of the seven views for $\mathcal L_{\mathrm{perc}}$.

Late in the rollout in \figref{fig:10}, the perceptual term reduces taillight bloom and ghosting while keeping body text, road markings, and small-object contours. This is evidence for those local failure modes, not a general frequency-response claim about DMD.

\subsection{RigCritic: cross-view adversarial refinement}\label{sec:4.5}

\Secrefs{sec:4.3}{sec:4.4} yield a stable one-step causal generator, but reverse KL $D_{\mathrm{KL}}(q_\theta\|p_{\mathrm{teacher}})$ is an expectation under the student: student samples that the teacher assigns low probability are penalized, whereas teacher modes rarely visited by the student contribute little gradient, leaving a mode-seeking bias. Adversarial training restores real video as the reference~\citep{goodfellow2014gan}. DMD2 jointly optimizes distribution matching and an adversarial term~\citep{yin2024dmd,yin2024dmd2}. AAD-1 stabilizes an asymmetric GAN with ODE initialization and a short DMD warm-up, then stops DMD before motion collapse~\citep{li2026aad1}. Driving failures often span cameras---fisheye and pinhole motion can disagree, or covisible regions can drift---so per-view discrimination is insufficient. We therefore introduce RigCritic. Following the staged schedule and generator--discriminator information asymmetry of~\citep{li2026aad1}, DMD updates end after \secref{sec:4.3}; the final stage optimizes only the adversarial objective together with the perceptual term of \secref{sec:4.4}. The discriminator scores the full latent clip---four fisheye and three pinhole views---as one rig sample, allowing cross-view geometry and temporal consistency to contribute directly to the adversarial gradient. Readout layers are selected by an influence probe rather than copied from fixed indices.

The generator $G_\theta$ remains strictly causal. The discriminator $D_\psi$ is initialized from bidirectional Wan2.2 and uses APT's learnable-query head~\citep{lin2025apt} to pool full-clip, multi-view context; future information is therefore available only during training~\citep{lin2025apt,li2026aad1}. Instead of copying the fixed readout layers used by APT or AAD-1, an input-dependent influence probe replaces each block with the identity, measures the resulting cosine shift in the final state, and selects taps across depth for appearance, motion, and semantics.

Training clips are autoregressively unrolled with the bounded history of \secref{sec:4.6}. Real and generated clips are noised at a random level $\tau$; unlike the shared noise of AAD-1~\citep{li2026aad1}, independent noise samples yield $\mathbf z_\tau$ and $\hat{\mathbf z}_\tau$ around each distribution. The logistic objectives are:

\begin{align}
\mathcal L_{\mathrm{adv}}^{D}&=\mathbb E_{\mathbf z\sim p_{\mathrm{data}}}\!\left[\operatorname{softplus}\!\left(-D_\psi(\mathbf z_\tau,\tau,\mathbf c)\right)\right]+\mathbb E_{\hat{\mathbf z}\sim G_\theta}\!\left[\operatorname{softplus}\!\left(D_\psi(\hat{\mathbf z}_\tau,\tau,\mathbf c)\right)\right],\label{eq:adv-d}\\
\mathcal L_{\mathrm{adv}}^{G}&=\mathbb E_{\hat{\mathbf z}\sim G_\theta}\!\left[\operatorname{softplus}\!\left(-D_\psi(\hat{\mathbf z}_\tau,\tau,\mathbf c)\right)\right].\label{eq:adv-g}
\end{align}

The discriminator also applies finite-difference R1/R2 on real and generated samples, with objective $\mathcal L_{\mathrm{adv}}^{D}+\lambda_R(\mathcal L_{R_1}+\mathcal L_{R_2})$, where $\lambda_R$ is the gradient-penalty weight. Finite-difference R1 comes from APT~\citep{lin2025apt}; symmetric R1/R2 follows AAPT and AAD-1~\citep{li2026aad1,lin2025aapt}.

Adversarial matching does not enforce local correspondence to paired ground truth, so RigCritic retains the \secref{sec:4.4} perceptual objective:

\begin{equation}
\mathcal L_G=\mathcal L_{\mathrm{adv}}^{G}+\lambda_{\mathrm{perc}}\mathcal L_{\mathrm{perc}}.
\label{eq:gen-adv}
\end{equation}

The discriminator contributes clip-level gradients from real videos, while the perceptual loss anchors paired roads, text, and small objects. Both objectives are used only during training; deployment remains one-step, causal, and bounded in history. This completes the post-training sequence in \Secrefs{sec:4.1}{sec:4.5}; \secref{sec:4.6} describes its long-horizon runtime.

\subsection{Long-horizon inference and a bounded KV-cache}\label{sec:4.6}

Closed-loop inference proceeds recursively over latent timesteps: predict the next seven-view latent, append it to history, and continue. There is no additional long-horizon training stage. Self-rollout (\secref{sec:4.3}) and clip-level discrimination (\secref{sec:4.5}) reduce the train--test history-source gap within the training window but do not expose the model to arbitrary sequence lengths. At inference, seven fixed-capacity KV-caches keep per-step memory independent of rollout duration. Within the training window the history is a FIFO of length $T$:

\begin{equation}
\mathbf z^{1:V}_{1:T}\leftarrow\big[\mathbf z^{1:V}_{2:T},\,\hat{\mathbf z}^{1:V}_{T+1}\big].
\label{eq:fifo}
\end{equation}

\textbf{Training versus inference windows.} The causal 4$\times$ VAE maps 81 RGB frames to 21 latents ($\approx$8.1 s at 10 FPS) and 301 frames to 76 latents ($\approx$30 s). All AR post-training phases stop at $T=21$: TF and CD (\secrefs{sec:4.1}{sec:4.2}) use full-clip teacher-forced forwards without a KV-cache, and the DMD self-rollout depth in \secref{sec:4.3} is also 21. The recurrence \eqrefn{eq:fifo} describes that training-window FIFO. The 30 s result reuses the same per-step loop with a rolling cache of capacity $W\ll76$; it therefore measures extrapolation beyond the training window, not long-window training. Because per-step memory does not grow with duration, the identical loop continues at minute scale; we report 30 s as the documented qualitative horizon.

\textbf{Bounded KV-cache.} Each view retains an $S$-latent attention-sink prefix and the most recent $W-S$ latents (e.g., $W=7$, $S=3$), evicting the oldest non-sink latent~\citep{xiao2024streamingllm}. Heterogeneous views advance independently at token size $h_vw_v$. RoPE uses global latent indices: eviction removes KV content without re-indexing time, unlike StreamingLLM's in-cache re-indexing~\citep{xiao2024streamingllm}. Cross-view attention reads only the current timestep's hidden states, not other views' caches. The sink prefix---including the start-of-clip latent---is a video-specific scene anchor; that use is ours, not a claim of~\citep{xiao2024streamingllm}.

\textbf{Condition injection.} Each step writes clean context at diffusion $t=0$ without direct control embeddings, injects control only while denoising the current latent, and then replaces that cache slot with the clean prediction. Historical KV therefore stores clean context without explicit control features. Layout is sliced to the target length; missing timestamps are represented by an all-zero layout, matching the no-constraint pattern used during training. Camera poses must be available at every requested timestamp.

\section{Training procedure and systems engineering}\label{sec:5}

\Secrefs{sec:3}{sec:4} define the heterogeneous multi-view architecture and autoregressive post-training objectives. This section explains how we train these models on a production rig within a single-node memory budget. The scheduler (\secref{sec:5.1}) treats each resolution--view-subset pair as an atomic sampling configuration, allowing full-rig topology coverage and high-resolution detail to be allocated independently. The curriculum (\secref{sec:5.2}) progresses from three-view low resolution to a mixture of seven-view low resolution and three-view high resolution, followed by cosine learning-rate annealing. Our training framework (\secref{sec:5.3}) makes these mixed configurations---and, separately, seven-view high-resolution autoregressive post-training---tractable. Scheduling determines the cost composition of each step, the curriculum determines its order, and systems optimization determines the feasible envelope.

\subsection{Heterogeneous mixed training}\label{sec:5.1}

A production camera system simultaneously has many views, unequal resolutions, and sparse effective covisibility. Always training at full-view high resolution on long sequences makes compute grow rapidly with spatiotemporal tokens; only lowering resolution loses distant agents, lane boundaries and fine motion cues needed for closed-loop driving. ZYT-World therefore decouples global geometric coverage from local detail: after entering the multi-view mixed stage, the low-resolution branch covers all seven cameras in the corresponding updates, while the high-resolution branch activates only strongly covisible view subsets. Both share backbone, camera semantic slots and CVA parameters, so raising resolution or changing the view combination does not require copying the model or redefining the network.

In the 7-view mixed stage the scheduler samples ``resolution configuration $\times$ view subset'' as the atomic unit, alternating 1:1 between a low-resolution full-rig stream and a high-resolution subset stream. Five three-view combinations cover all 12 edges of the seven-camera covisibility graph (left of submodule 5 in \figref{fig:6}: front cluster, side-bridge cluster, rear cluster), with the full seven-view combination retained; the front combination most relevant to driving decisions is up-weighted. Intra-configuration bucketing keeps sequence shape and attention masks static inside a batch; random interleaving across configurations prevents one view or resolution from dominating. When the high-resolution branch uses a shorter temporal extent we take the temporal prefix of the sample (first 41 of 81 frames) and rebuild camera and layout conditions with the same first frame as reference, so that the two streams stay strictly aligned on the prefix and the condition coordinate frame does not drift with the resolution configuration.

\sbox{\sidebox}{\begin{minipage}{0.50\columnwidth}
\figref{fig:11} reports steady-state cost under the sampling schedule, normalized to pretraining Stage 1 (3-view, 81-frame low resolution). Mixed configurations include 1:1 low/high alternation, and 41-frame high-resolution steps retain random 3-view/7-view subset sampling by training weight. Counting this schedule, the 41-frame mixed configuration costs $1.93\times$ a pretraining Stage-1 step on average; extending high-resolution context to 81 frames and fixing a 3-view subset raises the average to $2.66\times$.
\end{minipage}}
\noindent
\usebox{\sidebox}\hfill
\begin{minipage}[c][\sideboxheight][s]{0.47\columnwidth}
\centering
\includegraphics[width=\linewidth]{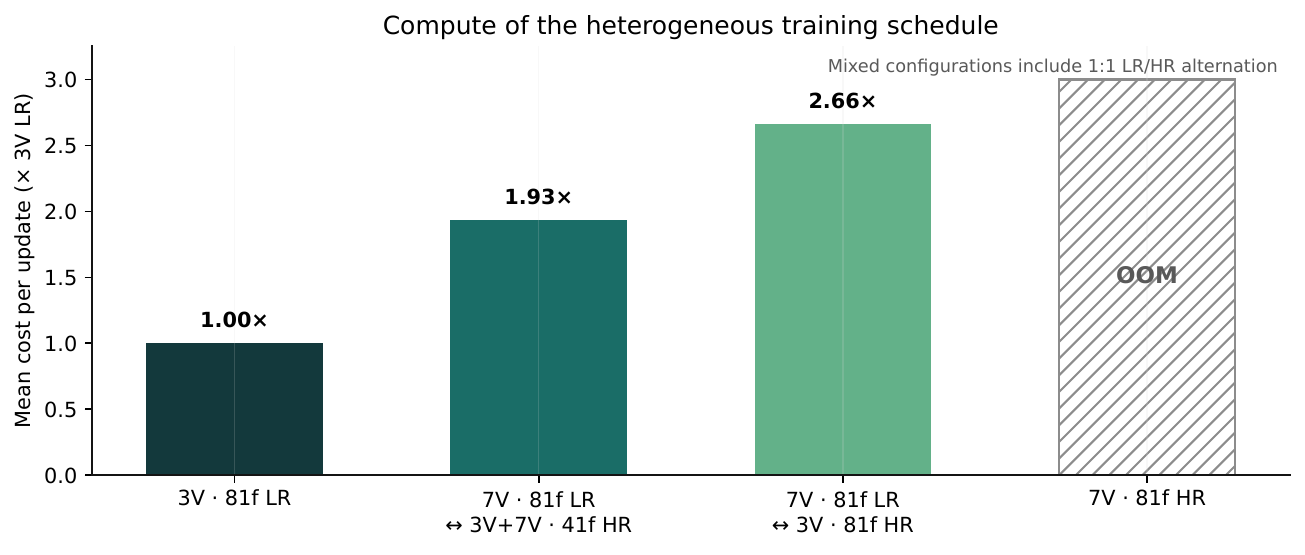}
\vfill
\captionof{figure}{Average compute of heterogeneous training, normalized to a 3-view low-resolution update.}\label{fig:11}
\end{minipage}

Direct 7-view, 81-frame high-resolution training exceeds the single-machine memory budget, so it has no comparable measured step cost. Heterogeneous scheduling therefore does more than shorten high-resolution sequences: it spends half of the updates at low resolution to keep full camera-topology coverage, and confines high-resolution compute to covisible subgraphs, gradually increasing spatiotemporal supervision inside a trainable memory envelope.

\subsection{Staged curriculum}\label{sec:5.2}

Pretraining follows a four-stage curriculum under a shared parameterization, with no architectural or weight remapping between stages. These pretraining Stages 1--4 are a data-and-resolution schedule and are not the TF/CD/DMD/RigCritic phases of \secref{sec:4}. Pretraining Stage 1 trains front wide, front narrow, and front fisheye at low resolution while progressively introducing the Pl\"ucker-ray, temporal, and conditioning modules. Stage 2 adds seven-view, 81-frame low-resolution sequences and alternates them 1:1 with randomly sampled 3-view or 7-view, 41-frame high-resolution subsets. Stage 3 replaces the high-resolution branch with a curated 3-view, 81-frame subset, concentrating long-horizon, high-fidelity supervision on the front wide, front narrow, and front fisheye combination most relevant to driving decisions. Stage 4 retains the Stage 3 data mixture and cosine-anneals the learning rate from $10^{-5}$ to $10^{-6}$.

\begin{figure}[H]
\centering
\includegraphics[width=0.98\textwidth,height=0.52\textheight,keepaspectratio]{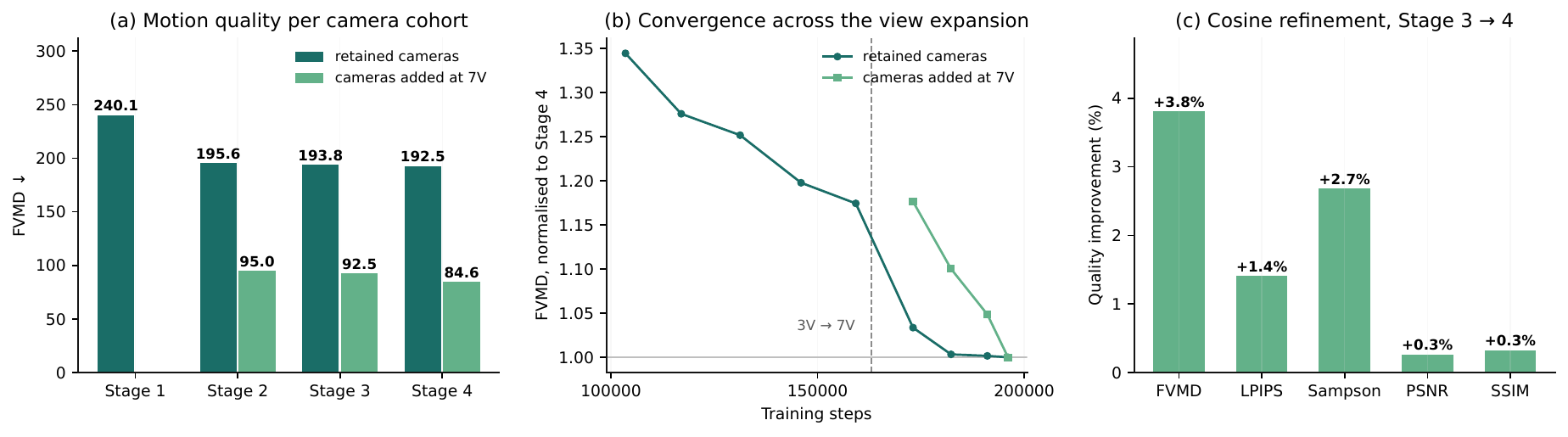}
\vspace{-1mm}
\caption{Pretraining-curriculum quality. (a) FVMD of already-trained vs newly merged cameras; (b) FVMD normalized to the pretraining Stage-4 final checkpoint; (c) relative improvement from the Stage-3 mean to the Stage-4 final checkpoint.}\label{fig:12}
\end{figure}

\figref{fig:12} tracks a continuous pretraining trajectory with FVMD (lower is better). Cameras present since Stage 1 improve from 240.1 to 192.5 by Stage 4, indicating that rig expansion does not regress the original views. Cameras introduced in Stage 2 improve from 95.0 to 84.6; the two groups are not directly comparable in absolute value. Under seven-view evaluation, FVMD decreases from the Stage-2 mean of 123.0 to 116.5 at the final Stage 4 checkpoint ($-$5.27\%). Stage 3 adds longer high-resolution context and a curated subset, whereas Stage 4 retains the data mixture and only anneals the learning rate; from the Stage-3 mean to that checkpoint, FVMD, LPIPS, Sampson, PSNR, and SSIM improve by 3.8\%, 1.4\%, 2.7\%, 0.3\%, and 0.3\%. The trajectory supports learning the core spatiotemporal prior at low cost in the few-view, low-resolution regime before adding views and resolution without changing the architecture.

\subsection{Distributed training and efficiency engineering}\label{sec:5.3}

All optimizations in this subsection are implemented in our distributed training framework built on Megatron-Core~\citep{shoeybi2019megatron}. The framework inherits TP, PP, SP, and DP; adds FSDP/HSDP; and rewrites scheduling, communication, and memory management to fit the workload of multi-view video diffusion. \Secrefs{sec:5.3.1}{sec:5.3.4} apply to both heterogeneous pretraining and autoregressive post-training, whereas Reuse Rollout, No-Reshard, and Segmented Backward are specific to the latter.

\subsubsection{Parallelism strategy}\label{sec:5.3.1}

Training composes HSDP~\citep{zhao2023fsdp} with DeepSpeed-Ulysses all-to-all sequence parallelism~\citep{jacobs2023ulysses}. HSDP shards parameters, gradients, and optimizer state within high-bandwidth groups and replicates them across groups; Ulysses redistributes activations between the sequence and attention-head dimensions. The corresponding process group is named SP in our framework, distinct from Megatron's tensor-parallel sequence parallelism. Replicating KV heads when their count is smaller than or not divisible by the sequence-parallel degree (as in GQA/MQA) is our layout adaptation.

\begin{figure}[!ht]
\centering
\includegraphics[width=0.98\textwidth,height=0.52\textheight,keepaspectratio]{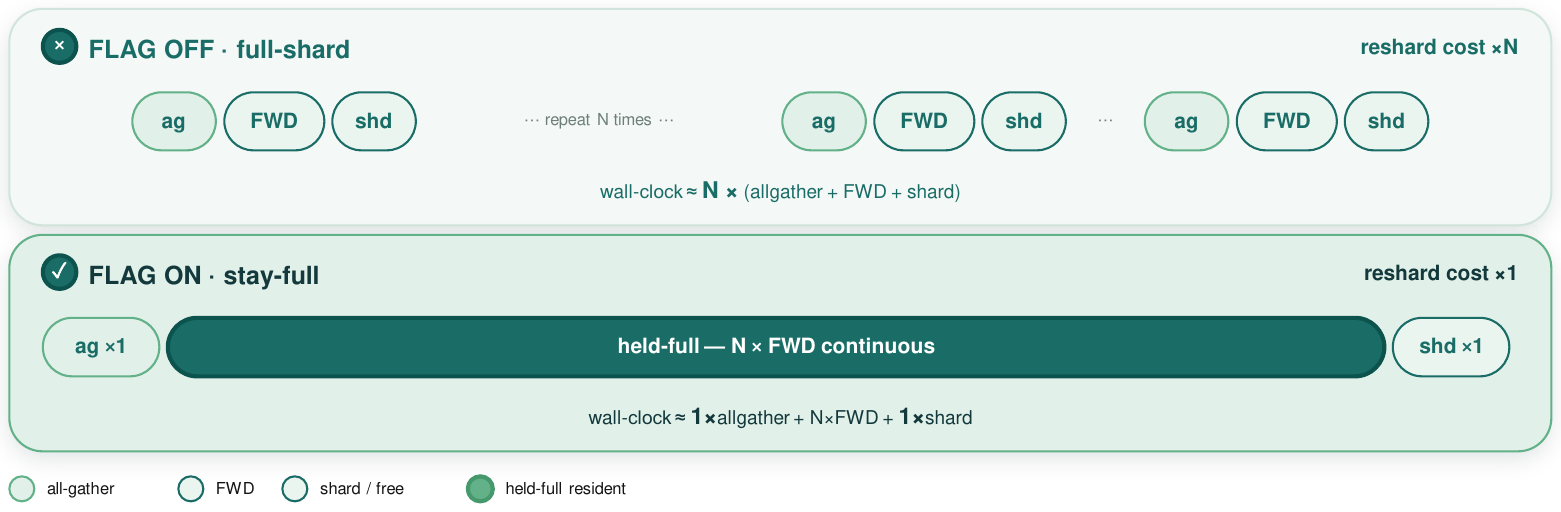}
\vspace{-1mm}
\caption{FSDP No-Reshard: default per-forward reshard (top) versus keeping parameters unsharded (bottom).}\label{fig:13}
\end{figure}

\begin{figure}[!ht]
\centering
\includegraphics[width=0.96\textwidth,height=0.62\textheight,keepaspectratio]{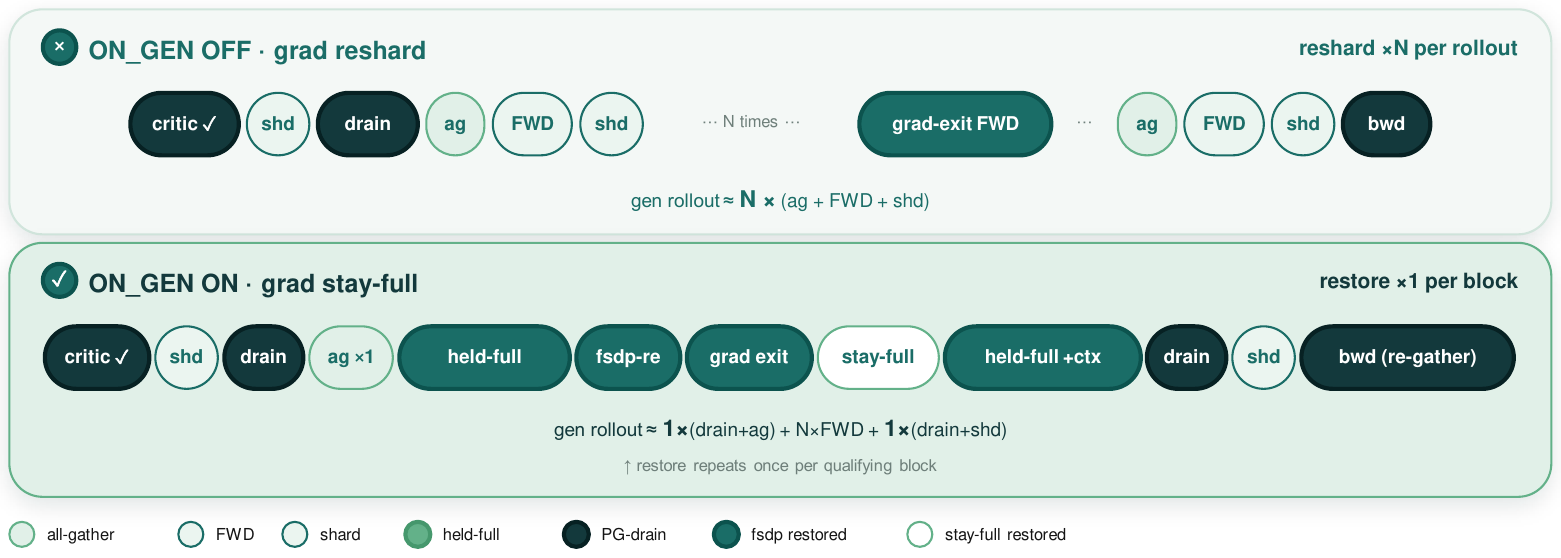}
\vspace{-1mm}
\caption{Default reshard versus No-Reshard on the DMD generator rollout.}\label{fig:14}
\end{figure}

\subsubsection{Compute optimization}\label{sec:5.3.2}

\textbf{Frame-local memory attention.} Each rank computes only its local frame's block-diagonal memory attention, eliminating the full-query all-gather and redundant work. At SP=8, layer compute and peak query memory fall to \textasciitilde{}1/8, and memory-training throughput rises \textasciitilde{}52\%.

\textbf{Reuse Rollout.} Detached generator rollouts are cached in memory released after generator backpropagation and reused by the DMD fake score model without increasing peak memory, halving its forward-pass time. Batching views reduces the rollout segment to approximately one sixth of its original cost.

\textbf{Condition rewrite.} Batched, sequence-parallel Pl\"ucker/layout encoding and blocked-matmul convolution reduce condition time by 88.7\% and end-to-end step time by \textasciitilde{}10\%.

\textbf{FlexAttention tuning.} Jointly tuned forward/backward tiles, warps, stages, and mask blocks improve forward/backward by 33.4\%/22.2\% without changing attention semantics.

\textbf{Memory-bound fusion.} Fusing AdaLN, RoPE, normalization, and casts avoids repeated long-sequence traffic and cuts small-op hot-path share from \textasciitilde{}26\% to below 10\%.

\subsubsection{Communication optimization}\label{sec:5.3.3}

Repeated FSDP all-gather accounted for 26.3\% of rollout latency. No-Reshard keeps full parameters across the entire rollout, reducing aggregation from $N$ times to one; backward restores standard sharding, and boundary synchronization fixes HSDP/sequence-parallel collective order (\figrefs{fig:13}{fig:14}).

\subsubsection{Memory optimization}\label{sec:5.3.4}

\textbf{Adaptive rematerialization.} The system selects the minimum rematerialization depth from the current resolution, duration, and view subset, separately balancing backbone, cross-view, and memory attention under one memory budget.

\textbf{Async offload.} Selected boundary activations move to host memory after forward and prefetch before backward, overlapping transfer with compute and saving \textasciitilde{}30 GB in the full configuration.

\textbf{Segmented backward.} Cutting the graph at the generated latents, we first backpropagate through the fake score model and perceptual loss, and then through the generator. Peak memory decreases from $M_G+M_{\mathrm{score}}+M_{\mathrm{perc}}$ to $\max(M_G,M_{\mathrm{score}}+M_{\mathrm{perc}})$, removing approximately 10 GB of graph co-residency while preserving parameter gradients.

\subsubsection{End-to-end gains and equivalence boundaries}\label{sec:5.3.5}

These optimizations make seven-view high-resolution autoregressive post-training feasible; the same configuration previously ran out of device memory. Once memory is no longer the bottleneck, the sequence-parallel degree can drop from 8 to 4, yielding 2.7$\times$--6.3$\times$ higher throughput at the same number of devices and global batch size.

\begin{figure}[!ht]
\centering
\includegraphics[width=0.98\textwidth,height=0.52\textheight,keepaspectratio]{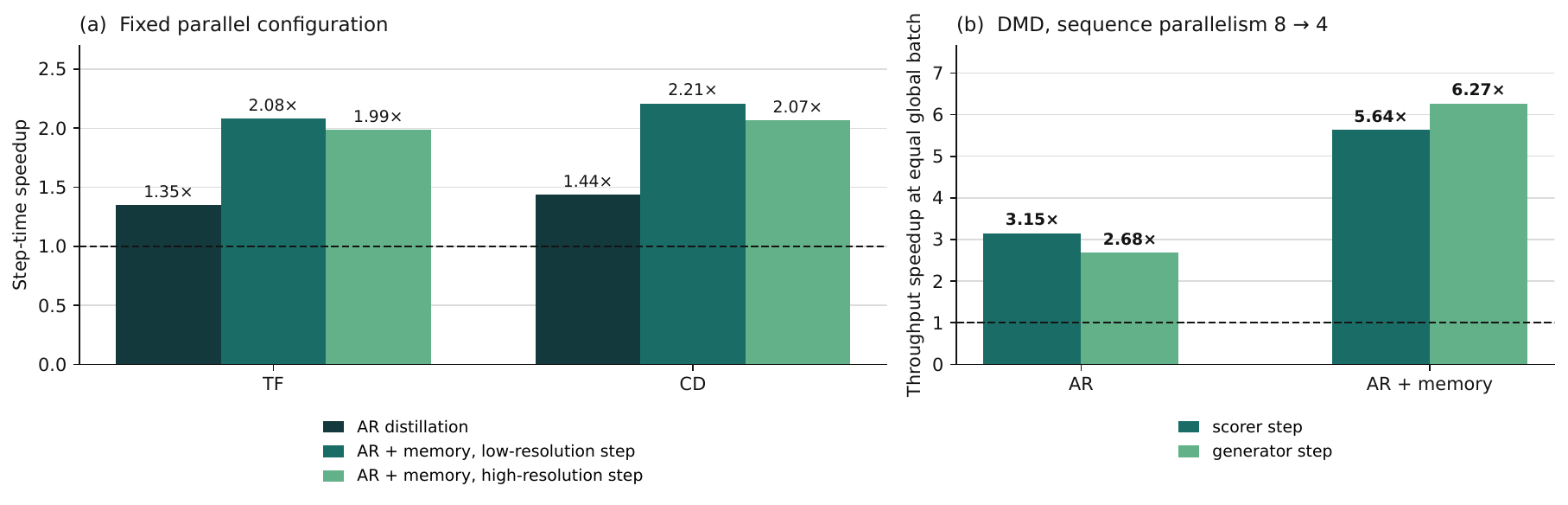}
\vspace{-1mm}
\caption{Training-system speedups for AR post-training. (a) Step time at a fixed parallel configuration for TF and CD; (b) DMD throughput after sequence parallelism $8\rightarrow4$. Bars compare the AR configuration with and without the memory path. Dashed line: pre-optimization baseline.}\label{fig:15}
\end{figure}

\figref{fig:15} reports end-to-end gains for AR post-training (TF, CD, DMD), not the pretraining Stages of \secref{sec:5.2}. TF and CD improve by 1.35$\times$ and 1.44$\times$ at a fixed parallel configuration; with memory optimizations, low- and high-resolution steps reach 2.08$\times$--2.21$\times$ and 1.99$\times$--2.07$\times$. After SP$8\rightarrow$SP$4$, DMD scorer and generator steps improve from 3.15$\times$/2.68$\times$ to 5.64$\times$/6.27$\times$. Combined optimizations accelerate bidirectional pretraining by approximately 1.65$\times$.

Frame-local execution, condition encoding, and segmented backward preserve the underlying computation; rematerialization and offload alter only tensor residency. Fused kernels are validated within numerical tolerance, while No-Reshard and Reuse Rollout are checked for collective ordering and cross-GPU gradient consistency. Because local speedups are not additive, \figref{fig:15} reports the end-to-end AR-phase gains.

\section{Real-time decoding and inference}\label{sec:6}

Closed-loop deployment requires optimizing the entire latent-to-sensor path, not generation quality alone. TinyVAE reduces pixel-decoding cost; few-step post-training reduces backbone evaluations per latent timestep; W8A8 reduces the cost of each matrix multiplication; and our inference engine fuses operators, reuses temporal state, and schedules multi-view execution. These components target decoding, the number of sampling steps, numerical precision, and execution efficiency, respectively.

\subsection{The TinyVAE decoder}\label{sec:6.1}

The latency of the Wan decoder can hardly meet the real-time requirement of closed-loop simulation, so the decoding stage must be accelerated. TinyVAE must therefore approach the decoding quality of Wan in both pixel space and semantic space at a small fraction of the latency and GPU memory.

TinyVAE has 19M parameters, less than 4\% of the 555M-parameter Wan decoder. The Wan encoder is kept frozen and only the decoder is trained, so training reduces to deterministic regression from latents to pixels. We start from TAEHV super, the Wan-compatible release of TAEHV~\citep{boerbohan2025taehv}, which serves as the baseline.

\textbf{Notation.} Let $\hat{\mathbf x}=D_{\mathrm{light}}(\mathbf z)$ denote the clip decoded from the latent $\mathbf z$ and $\mathbf x$ the paired RGB clip, both with pixel values in $[0,1]$. A training batch contains $N$ frames indexed by $n\in\{1,\dots,N\}$, with $\hat{\mathbf x}_n,\mathbf x_n\in[0,1]^{3\times H\times W}$. The color channel is indexed by $c\in\{1,2,3\}$ and the pixel position by $(h,w)\in\{0,\dots,H-1\}\times\{0,\dots,W-1\}$, so that $x_{n,c,h,w}$ and $\hat x_{n,c,h,w}$ denote the entries of $\mathbf x_n$ and $\hat{\mathbf x}_n$.

\textbf{Objective.} The decoder minimizes
\begin{equation}
\mathcal L_{\mathrm{dec}}
=\alpha\,\mathcal L_{1}
+\lambda_{\mathrm p}\,\mathcal L_{\mathrm{LPIPS}}
+\lambda_{\mathrm d}\,\mathcal L_{\mathrm{DINO}}
+\lambda_{\mathrm g}\,\mathcal L_{\mathrm{grid}},
\label{eq:tinyvae}
\end{equation}
with $(\alpha,\lambda_{\mathrm p},\lambda_{\mathrm d},\lambda_{\mathrm g})=(0.1,\,0.1,\,0.02,\,1)$. The four terms address pixel fidelity, perceptual similarity, semantic consistency, and lattice-aligned residuals, respectively.

\textbf{Pixel term.} The pixel term is the mean absolute error over frames, channels, and pixels,
\begin{equation}
\mathcal L_{1}=\frac{1}{3NHW}\sum_{n,c,h,w}\bigl|\hat x_{n,c,h,w}-x_{n,c,h,w}\bigr|.
\label{eq:l1}
\end{equation}
We use the L1 norm rather than MSE because, at matched gradient scale, MSE leaves stronger periodic residuals on smooth regions.

\textbf{Perceptual term.} Let $d_{\mathrm{LPIPS}}(\cdot,\cdot)$ denote the frozen LPIPS distance with AlexNet features~\citep{zhang2018lpips}, evaluated on inputs mapped from $[0,1]$ to $[-1,1]$ by $x\mapsto 2x-1$. The perceptual term is the mean over frames,
\begin{equation}
\mathcal L_{\mathrm{LPIPS}}=\frac1N\sum_{n=1}^{N} d_{\mathrm{LPIPS}}\bigl(\hat{\mathbf x}_n,\mathbf x_n\bigr).
\label{eq:lpips}
\end{equation}
This feature-space distance is correlated with human similarity judgments; without it the decoder converges toward a blurred pixel-average solution.

\textbf{Semantic term.} Let $\phi_\ell(\mathbf x_n)\in\mathbb R^{K\times C}$ denote the patch tokens (excluding the class token) output by Transformer block $\ell$ of a frozen DINOv2 ViT-B/14~\citep{oquab2023dinov2} and passed through the network's final LayerNorm, where $K$ is the number of patch tokens and $C$ the embedding dimension. Each frame is bilinearly resized to $224\times560$ and ImageNet-normalized before encoding, giving $K=16\times40=640$ and $C=768$. Let $\mathcal S=\{3,6,9,12\}$ be the set of supervised blocks and $\lVert\cdot\rVert_{\mathrm F}$ the Frobenius norm. The semantic term is
\begin{equation}
\mathcal L_{\mathrm{DINO}}
=\frac{1}{N\,|\mathcal S|}\sum_{n=1}^{N}\sum_{\ell\in\mathcal S}
\frac{\lVert\phi_\ell(\hat{\mathbf x}_n)-\phi_\ell(\mathbf x_n)\rVert_{\mathrm F}^2}{\lVert\phi_\ell(\mathbf x_n)\rVert_{\mathrm F}^2},
\label{eq:dino}
\end{equation}
where the target tokens $\phi_\ell(\mathbf x_n)$ are treated as constants. Normalizing each block by the energy of its target tokens equalizes the contributions of the four depths, and the mid-level blocks retain a sensitivity to structure and texture that the final block abstracts away. The term acts as a semantic regularizer: it leaves reconstruction PSNR/SSIM essentially unchanged while improving the distributional quality of generated frames.

\textbf{Lattice term.} Let $\mathbf r_n=\hat{\mathbf x}_n-\mathbf x_n$ be the residual of frame $n$ with entries $r_{n,c,h,w}$, spatially cropped to $H'\times W'$, where $H'$ and $W'$ are the largest multiples of the lattice period $P=16$ not exceeding $H$ and $W$. Each pixel $(h,w)$ is assigned the lattice phase $(i,j)=(h\bmod P,\,w\bmod P)$, and the residual is averaged over all pixels sharing the same phase:
\begin{equation}
m_{n,c,i,j}=\frac{P^2}{H'W'}\sum_{u=0}^{H'/P-1}\sum_{v=0}^{W'/P-1} r_{n,c,\,uP+i,\,vP+j},
\qquad i,j\in\{0,\dots,P-1\}.
\label{eq:lattice-mean}
\end{equation}
The phase-wise average preserves any residual component whose spatial period divides $P$, whereas residual content not aligned with the lattice is attenuated. Subtracting the per-frame, per-channel mean over the $P^2$ phases, $\bar m_{n,c}=P^{-2}\sum_{i,j}m_{n,c,i,j}$, removes the constant offset already penalized by $\mathcal L_1$ and retains only the periodic component. The lattice term is the L1 norm of the resulting phase-binned residual, averaged over frames, channels, and phases:
\begin{equation}
\mathcal L_{\mathrm{grid}}
=\frac{1}{3NP^2}\sum_{n=1}^{N}\sum_{c=1}^{3}\sum_{i,j=0}^{P-1}\bigl|m_{n,c,i,j}-\bar m_{n,c}\bigr|.
\label{eq:grid}
\end{equation}
The L1 norm is used on the binned residual because the gradient of a squared norm vanishes as the residual amplitude tends to zero, so low-amplitude periodic residuals would not be removed. In practice, this term markedly suppresses checkerboard artifacts on smooth regions and also benefits perceptual and distributional quality.

\subsection{Few-step AR inference}\label{sec:6.2}

A few-step student generates synchronized seven-view latent timesteps rather than processing cameras sequentially or entire videos as clips. Given bounded history $\mathcal H_t$ (\secref{sec:4.6}) and control $\mathbf c_{t+1}$, each denoising substep updates all views jointly, so every generated latent corresponds to the same physical time interval.

For $K$ denoising substeps of a distilled student $\Phi_{\theta,k}$, one latent timestep is updated as

\begin{equation}
\mathbf z_{t+1}^{(0)}\sim\mathcal N(0,\mathbf I),\qquad \mathbf z_{t+1}^{(k+1)}=\Phi_{\theta,k}\!\left(\mathbf z_{t+1}^{(k)},\mathcal H_t,\mathbf c_{t+1}\right),\quad k=0,\ldots,K-1,
\label{eq:fewstep}
\end{equation}

where $\mathbf z_{t+1}$ stacks all seven views and $\mathcal H_t$ is the bounded KV history. The $K$ intra-timestep denoising updates read historical KV without modifying it; noisy intermediates never enter the cache. New keys and values are committed only after the final clean latent. This solve-then-commit rule keeps cache semantics consistent across the 1-, 2-, and 4-step students.

The 1/2/4-step operating points are separately distilled students, not early exits from one checkpoint. Generating $T$ latent timesteps costs $KT$ backbone forwards, while the fixed history window is independent of $T$. Camera, layout, and ego-motion encodings are reused across substeps of a timestep; its final latent enters the cache and can be decoded immediately by TinyVAE. Comparisons in \secref{sec:7.2.3} disable memory.

\subsection{W8A8 quantization}\label{sec:6.3}

Few-step distillation cuts the number of forwards; W8A8 reduces each DiT forward. In closed-loop rollout, visual and control tokens vary strongly in scale across tokens, while activation statistics drift over generated time; naive per-tensor quantization is further dominated by fixed post-normalization outlier channels. We address channel outliers, token-wise dynamic range, and closed-loop distribution shift with three complementary mechanisms.

SmoothQuant~\citep{xiao2023smoothquant} transfers activation outliers on $\mathbf X$ onto weights $\mathbf W$ through per-channel scales $\mathbf s$, with migration strength $\alpha_{\mathrm{sq}}\in[0,1]$:

\begin{equation}
\mathbf{Y}=(\mathbf{X}\,\mathrm{diag}(\mathbf{s})^{-1})(\mathrm{diag}(\mathbf{s})\mathbf{W}),\qquad s_j=\frac{\max(|\mathbf{X}_j|)^{\alpha_{\mathrm{sq}}}}{\max(|\mathbf{W}_j|)^{1-\alpha_{\mathrm{sq}}}}.
\label{eq:smoothquant}
\end{equation}

At the algorithmic level, channel scales are absorbed offline into the preceding Linear and AdaLN affine parameters, adding no extra matrix multiplication~\citep{xiao2023smoothquant}. When smoothing crosses a residual addition, the residual branch receives the same scaling to preserve algebraic equivalence.

Activations use ZeroQuant-style per-token dynamic quantization~\citep{yao2022zeroquant}, whereas weights use grouped static scales. Per-token scales absorb variation across token types and rollout time; grouped weight scales bound the error introduced after outlier migration. INT8 products accumulate in INT32. Let $\hat{\mathbf X}$ and $\hat{\mathbf W}$ be the smoothed activations and weights, $\Delta_{X,i}$ the per-token dequant scale, and $\Delta_{W,g}$ the scale of output-channel group $g$. In the inference-engine epilogue, $\Delta_X$ is applied per row (token) and $\Delta_W$ per group by broadcasting, not as a dense right multiplication:

\begin{equation}
\Delta_{X,i}=\frac{\max_j|\hat X_{i,j}|}{2^7-1},\qquad \Delta_{W,g}=\frac{\max_{(i,j)\in g}|\hat W_{i,j}|}{2^7-1},\qquad \mathbf{Y}\approx\mathrm{diag}(\Delta_X)(\mathbf{X}_{8}\mathbf{W}_{8})\,\Delta_W.
\label{eq:w8a8}
\end{equation}

Here $\mathbf{X}_{8}$ and $\mathbf{W}_{8}$ are the INT8 tensors. Channel-outlier migration follows SmoothQuant~\citep{xiao2023smoothquant} and is not part of the original ZeroQuant recipe.

The three mechanisms address different error sources: per-token scaling adapts to variation across token types and rollout time; SmoothQuant handles persistent channel outliers along the reduction dimension; and grouped weight scales limit the error introduced by outlier migration. Calibration emphasizes self-rollouts, records statistics separately for visual, camera, and ego-motion tokens, and merges them using a robust quantile; online activation scales absorb residual temporal drift. Softmax, normalization statistics, and positional encoding remain in FP16/BF16 or FP32. Our engine implements the resulting INT8 GEMM and fused epilogue.

\subsection{Inference engine}\label{sec:6.4}

Seven-view autoregression multiplies kernel-launch and memory-traffic overhead across Transformer depth. The engine integrates W8A8 GEMM, intra-block fusion, cross-view scheduling, temporal-state reuse, and sequence parallelism into a static incremental execution path. A single GPU is the primary operating point; when the context exceeds its capacity, the same execution semantics extend across two GPUs along the sequence dimension.

\textbf{Fused low-bit execution.} Weights are quantized offline and activations dynamically per token. Residual SmoothQuant scales that cannot be fully absorbed into Linear/AdaLN are applied inside the engine's INT8 quantization kernel; per-column bias and dequantization scales are fused into the GEMM epilogue. BF16 rounding points match the unfused baseline to limit numerical drift.

\textbf{Fusion of memory-bound operators.} Across 30 Transformer blocks, the engine fuses AdaLN addition and casting, replaces eight FP32 RoPE elementwise operations with one kernel that reuses Q/K sinusoids, fuses gated residuals, and removes redundant attention-path casts, avoiding large FP32 intermediates.

\textbf{Two-GPU incremental sequence parallelism.} We extend the training-time DeepSpeed-Ulysses all-to-all layout~\citep{jacobs2023ulysses} to inference: weights are replicated, tokens are sequence-sharded, and attention alternates between sequence and head sharding. KV-caches remain head-sharded; each denoising step exchanges only the current timestep's short QKV, so communication does not grow with accumulated context length. Incremental-cache sharding, communication, and state management are inference-side extensions of~\citep{jacobs2023ulysses}.

Correctness is validated hierarchically: low-bit and fused kernels against unfused layer outputs, incremental caching against full-sequence forwards, covisibility sparsity against dense attention, and complete rollouts for NaN/Inf and accumulated error. Local speedups are not additive; \secref{sec:7.2.3} reports the end-to-end few-step generator operating points. Taken together, our engine on two GPUs reaches a \best{2.72$\times$} speedup---quantization and kernel fusion contribute \best{1.6$\times$}, dual-GPU sequence parallelism a further \best{1.7$\times$} after communication overhead---and is the 4 FPS seven-view operating point in \figref{fig:1}.

\section{Evaluation and results}\label{sec:7}

We evaluate ZYT-World as a sensor generator for closed-loop driving simulation: whether videos keep appearance, temporal coherence, controllable content, and cross-view geometry, and whether latency meets the target. All numbers here are open-loop video, detector, geometry, and systems measurements on an internal split.

\subsection{Evaluation protocol}\label{sec:7.1}

The protocol is organized around four simulator failure modes. Appearance fidelity is measured with frame-level FID~\citep{heusel2017ttur} and FD-DINOv2 (FDD)~\citep{stein2023exposing}, together with paired PSNR, SSIM~\citep{wang2004ssim}, and LPIPS~\citep{zhang2018lpips}; PSNR is interpreted only under fixed content and distortion type~\citep{huynhthu2008psnr}. Temporal coherence is measured with video-level FVD~\citep{unterthiner2018fvd} and keypoint velocity- and acceleration-based FVMD~\citep{liu2024fvmd}. Content alignment to the corresponding real video is measured with YOLO Det-F1, paired IoU, true-positive IoU, and object-count error. Cross-view geometry is measured with epipolar Sampson error~\citep{hartley2004mvg}. Fixed-latent reconstruction comparisons additionally report \emph{character similarity per box} (char\_sim/box), since full-frame metrics do not necessarily reflect how faithfully small text is decoded. A fixed OCR engine detects and reads text boxes on the real frame (boxes it cannot reliably read there are discarded), then reads the same boxes on the decoded frame; each box scores $1-\mathrm{ED}(s_{\mathrm{dec}},s_{\mathrm{real}})/\max(|s_{\mathrm{real}}|,|s_{\mathrm{dec}}|)$ with $\mathrm{ED}$ the Levenshtein distance, averaged over boxes per clip and then over clips. These are diagnostic proxies for simulator suitability, not a substitute for policy-level closed-loop evaluation.

\begin{figure}[!ht]
\centering
\includegraphics[width=0.98\textwidth,height=0.52\textheight,keepaspectratio]{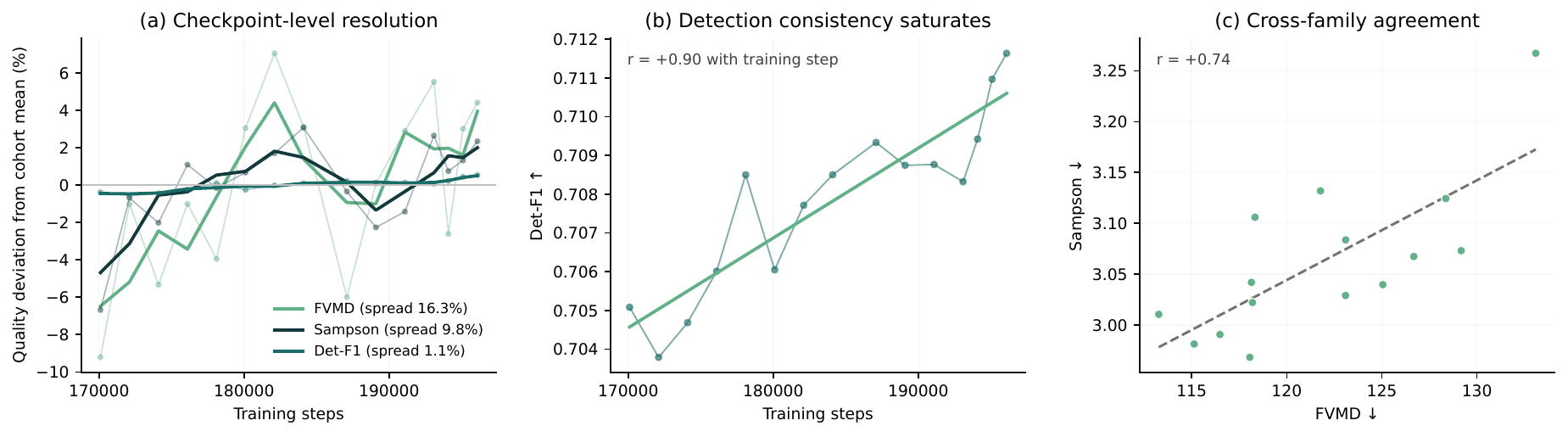}
\vspace{-1mm}
\caption{Checkpoint-level diagnostics on seven-view checkpoints. (a) Quality deviation from the cohort mean for FVMD, Sampson, and Det-F1 (faint: raw; thick: smoothed; spread in the legend); (b) Det-F1 versus training step; (c) FVMD versus Sampson.}\label{fig:16}
\end{figure}

Metric choice depends on the comparison. Across pretraining checkpoints, low-resolution loss correlates most strongly with FVMD and FVD, so temporal motion metrics are emphasized for convergence. \figref{fig:16} diagnoses how well three metrics separate neighbouring seven-view checkpoints, where the evaluation protocol is fixed so spread reflects the metric rather than a change of configuration. Panel (a) plots FVMD, Sampson, and Det-F1 as quality deviation from the cohort mean: the faint traces are raw checkpoints and the thick traces are three-point smoothed; the legend reports relative spread. FVMD moves over a $16.3\%$ range, Sampson $9.8\%$, and Det-F1 only $1.1\%$. Panel (b) shows that Det-F1 still trends with training step ($r=+0.90$) but saturates in a narrow band. Panel (c) shows that FVMD and Sampson co-vary ($r=+0.74$). Distillation compares the teacher and 1-, 2-, and 4-step students at a fixed 72-frame prediction horizon using FID, FDD, FVD, LPIPS, PSNR, and SSIM. Memory ablations use the same bidirectional 40-step sampler, so changes isolate the memory condition rather than the generation paradigm. TinyVAE is evaluated twice: decoding generated latents tests downstream generation quality, whereas fixed-latent reconstruction isolates decoder error.

Detection metrics use a fixed YOLO evaluator on generated and corresponding real videos, with matched classes and confidence thresholds. We report precision, recall, Det-F1, paired IoU, true-positive IoU, and count error; these quantify detector-level content agreement rather than human perceptual quality. FDD uses frozen DINOv2 features~\citep{stein2023exposing} and is kept distinct from pixel-domain metrics.

\subsection{Quantitative results}\label{sec:7.2}

We split results by what actually changed, rather than ranking different generation paradigms in one table. \secref{sec:7.2.1} is the bidirectional pretraining trajectory and the selected base; \secref{sec:7.2.2} holds the 40-step bidirectional sampler fixed and adds memory; \secref{sec:7.2.3} reports 1/2/4-step autoregressive students; \secref{sec:7.2.4} reports decoder generation quality, reconstruction fidelity, and efficiency. TinyVAE changes the latent-to-pixel channel, not the generator, so it is not ranked with the three generator groups. Loss design stays in \secref{sec:6.1}.

\subsubsection{Convergence of bidirectional pretraining}\label{sec:7.2.1}

\begin{figure}[!ht]
\centering
\includegraphics[width=0.96\textwidth,height=0.62\textheight,keepaspectratio]{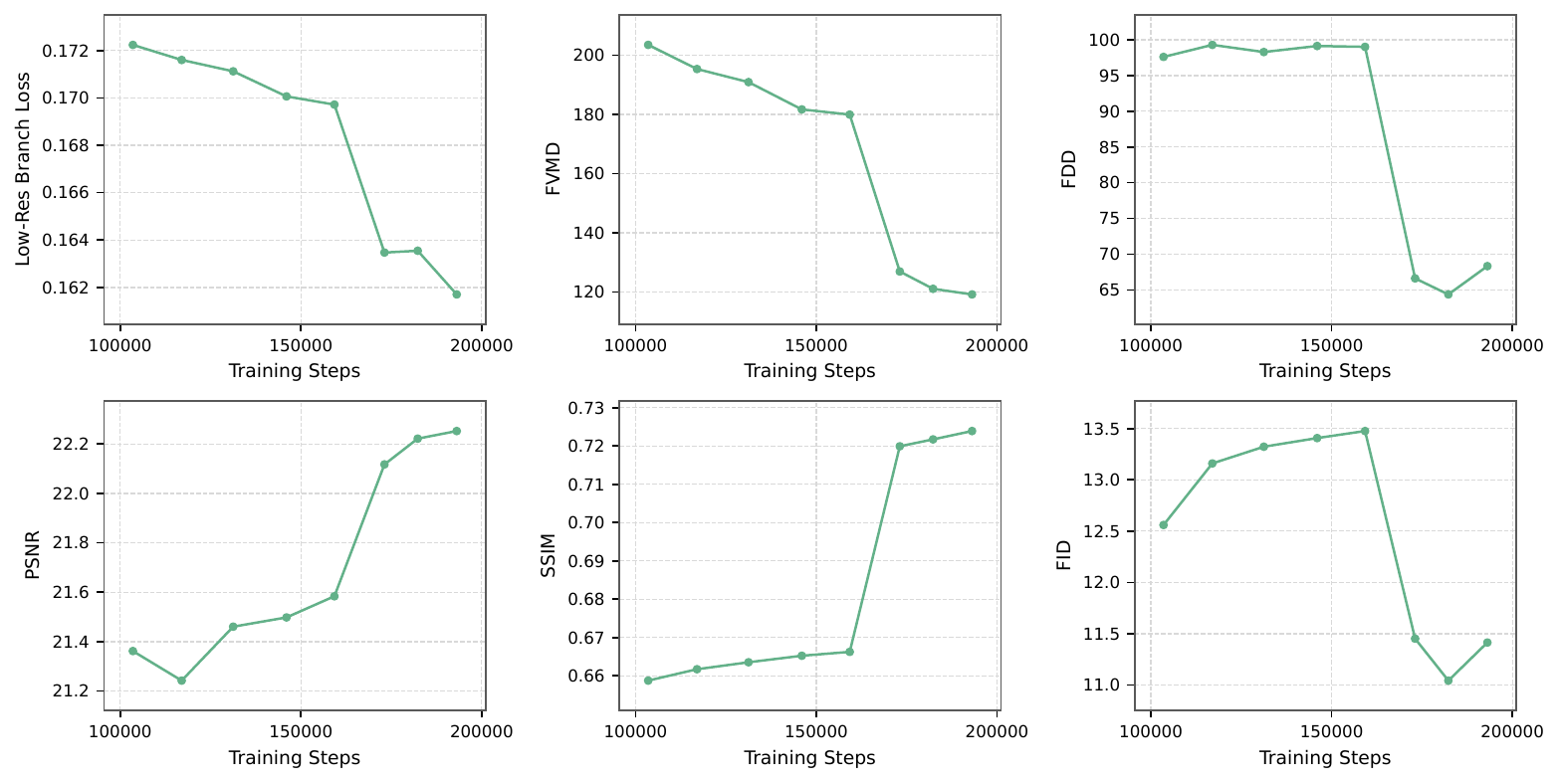}
\vspace{-1mm}
\caption{Pretraining Stage-4 trajectories of low-resolution loss, FVMD, FDD, PSNR, SSIM, and FID.}\label{fig:17}
\end{figure}

\figref{fig:17} traces the Stage 4 bidirectional teacher over cumulative training steps; \tabref{tab:2} reports the selected Stage 4 checkpoint without memory (\textit{w/o} Memory) and the memory-conditioned model (\textit{w/} Memory) under one evaluation protocol. The trajectory is for checkpoint selection on this run, not a universal map from training loss to downstream metrics.

\begin{table}[!ht]
\centering
\caption{Same-protocol evaluation of \textit{w/o} Memory versus \textit{w/} Memory.}\label{tab:2}
\scriptsize
\setlength{\tabcolsep}{3pt}
\renewcommand{\arraystretch}{1.08}
\resizebox{\textwidth}{!}{%
\begin{tabular}{l c c c c c c c c c }
\toprule
Model & FID$\downarrow$ & FVD$\downarrow$ & FVMD$\downarrow$ & FDD$\downarrow$ & LPIPS$\downarrow$ & PSNR$\uparrow$ & SSIM$\uparrow$ & Det-F1$\uparrow$ & Sampson$\downarrow$ \\
\midrule
\textit{w/o} Memory & \best{11.64} & 116.58 & 116.49 & 68.79 & 0.241 & 22.30 & 0.725 & 0.712 & 2.991 \\
\textbf{\textit{w/} Memory} & 11.72 & \best{113.63} & \best{101.93} & \best{64.55} & \best{0.216} & \best{23.06} & \best{0.741} & \best{0.726} & \best{2.965} \\
\bottomrule
\end{tabular}%
}
\end{table}

\subsubsection{Effect of the implicit memory condition}\label{sec:7.2.2}

\sbox{\sidebox}{\begin{minipage}[t]{0.355\columnwidth}
\vspace{0pt}\centering
\includegraphics[width=\linewidth]{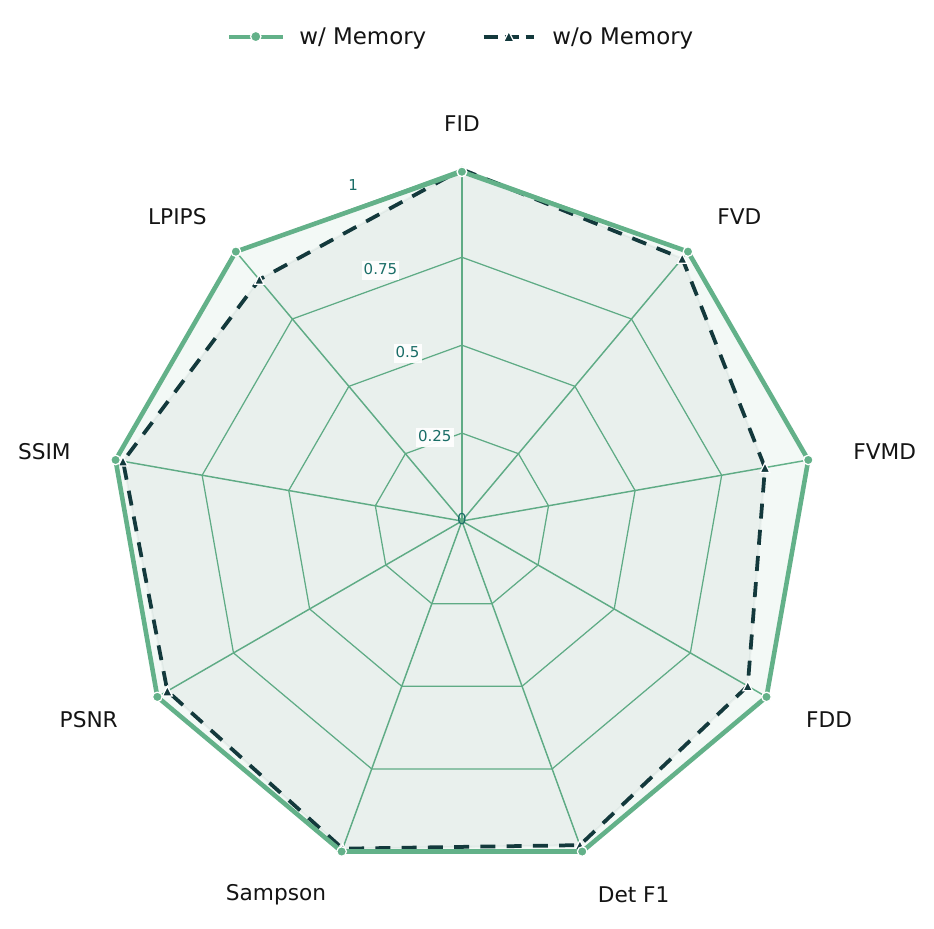}
\captionof{figure}{Capability envelopes of \textit{w/o} Memory versus \textit{w/} Memory.}\label{fig:18}
\end{minipage}}
\noindent
\begin{minipage}[t]{0.615\columnwidth}
\vspace{0pt}\tabref{tab:2} uses the same 249 scenes, 1,743 view streams, and 72-frame prediction horizon. \textit{w/} Memory is a separately trained checkpoint and the table reports a single test split without confidence intervals, so the observed differences are descriptive rather than a statistically isolated memory ablation.

\medskip
\figref{fig:18} plots the same comparison as outward-better capability envelopes, with \textit{w/o} Memory fixed at 1.0: FID, FVD, FVMD, FDD, LPIPS and Sampson use (\textit{w/o} Memory)/(\textit{w/} Memory), while PSNR, SSIM and Det-F1 use the reciprocal. Both models use the same target videos, prediction horizon and evaluation implementation.

\medskip
Aggregate quality remains comparable, indicating that adding the memory condition does not materially degrade the 40-step bidirectional generator; the memory path changes revisit behavior without trading away same-protocol generation quality.
\end{minipage}\hfill
\usebox{\sidebox}

\subsubsection{Distillation to few-step autoregressive generation}\label{sec:7.2.3}

\tabref{tab:3} compares three separately trained autoregressive operating points on 249 scenes and 1,743 view streams; it is not a sampling-step ablation of one checkpoint. Boldface marks the best among the three AR students only, following the column arrows (FID, FVD, FDD, LPIPS, Sampson: lower; speedup, PSNR, SSIM, Det-F1: higher); the teacher is a reference and is not ranked. Among the students, 2-step AR has the best FID, FVD, and LPIPS; 4-step AR has the best FDD, PSNR, and SSIM; 1-step AR has the best Det-F1, Sampson, and first-latent speedup. The teacher remains better on FDD, LPIPS, PSNR, SSIM, Det-F1, and Sampson; 2-step AR improves FID and FVD over the teacher. Speedup is first-latent latency: both the 40-step teacher and the AR student are timed on one latent timestep (\best{39.6$\times$}), not a full bidirectional clip against one AR unit. The separate timing setup in \figref{fig:2} gives the \best{107.7$\times$} generator-only speedup; the two numbers are not interchangeable, and neither is the 4 FPS seven-view inference-engine measurement in \figref{fig:1}.

\begin{table}[!tb]
\centering
\caption{Bidirectional teacher and separately trained few-step AR operating points. Boldface is the best AR student per column (teacher is a reference). Speedup is generator-side first-latent wall-clock versus the 40-step teacher on one latent timestep, excluding TinyVAE, W8A8, and our inference engine.}\label{tab:3}
\scriptsize
\setlength{\tabcolsep}{3pt}
\renewcommand{\arraystretch}{1.08}
\resizebox{\textwidth}{!}{%
\begin{tabular}{l c c c c c c c c c }
\toprule
Generator & Rel. speedup$\uparrow$ & FID$\downarrow$ & FVD$\downarrow$ & FDD$\downarrow$ & LPIPS$\downarrow$ & PSNR$\uparrow$ & SSIM$\uparrow$ & Det-F1$\uparrow$ & Sampson$\downarrow$ \\
\midrule
Bidirectional teacher & 1.0$\times$ (ref.) & 11.64 & 116.58 & 68.79 & 0.2411 & 22.30 & 0.7254 & 0.712 & 2.991 \\
4-step AR & 15.8$\times$ & 10.52 & 118.22 & \best{76.36} & 0.2677 & \best{21.27} & \best{0.6954} & 0.567 & 3.480 \\
2-step AR & 26.4$\times$ & \best{9.97} & \best{102.50} & 77.43 & \best{0.2646} & 21.26 & 0.6911 & 0.574 & 3.556 \\
1-step AR & \best{39.6$\times$} & 11.83 & 123.28 & 81.79 & 0.2655 & 21.07 & 0.6951 & \best{0.581} & \best{3.413} \\
\bottomrule
\end{tabular}%
}
\end{table}

\subsubsection{TinyVAE: generation, reconstruction, and efficiency}\label{sec:7.2.4}

The generator determines what information the latent carries; the decoder determines how faithfully it reaches pixel space. We therefore evaluate decoders in two settings that answer different questions. In \tabref{tab:4}, all three decoders receive identical latents produced by the bidirectional model, so differences reflect downstream generation quality. In \tabref{tab:5}, they decode latents obtained by encoding ground-truth clips with the frozen Wan encoder on the high-resolution reconstruction set (248 clips, 81 frames each), isolating decoder error from the generator. The compared decoders are Wan (555M parameters), the released Wan-compatible TAEHV weights (TAEHV super, 19M), and TinyVAE (19M, \secref{sec:6.1}). In both tables, boldface marks the best of the three decoders per column, following the column arrows.

\begin{table}[t]
\centering
\caption{Generation quality in the full generation chain. The same latents, produced by the bidirectional model, are decoded by each of the three decoders; only the decoder differs between rows.}\label{tab:4}
\scriptsize
\setlength{\tabcolsep}{3pt}
\renewcommand{\arraystretch}{1.08}
\resizebox{\textwidth}{!}{%
\begin{tabular}{l c c c c c c c c }
\toprule
Model & FID$\downarrow$ & FVD$\downarrow$ & FDD$\downarrow$ & LPIPS$\downarrow$ & PSNR$\uparrow$ & SSIM$\uparrow$ & Det-F1$\uparrow$ & Sampson$\downarrow$ \\
\midrule
Wan & 11.6415 & 116.5837 & \best{68.7932} & 0.2411 & 22.3026 & 0.7254 & \best{0.7116} & \best{2.9908} \\
TAEHV super & 12.8877 & 127.9301 & 86.8353 & 0.2487 & \best{22.4476} & \best{0.7280} & 0.6867 & 3.0697 \\
TinyVAE & \best{11.6306} & \best{115.9042} & 72.0696 & \best{0.2356} & 22.3884 & 0.7249 & 0.7053 & 3.0601 \\
\bottomrule
\end{tabular}%
}
\end{table}

\textbf{Generation.} On latents from the bidirectional model (\tabref{tab:4}), TinyVAE is better than TAEHV super on six of the eight metrics: FID 12.8877$\rightarrow$11.6306, FVD 127.9301$\rightarrow$115.9042, FDD 86.8353$\rightarrow$72.0696, LPIPS 0.2487$\rightarrow$0.2356, Det-F1 0.6867$\rightarrow$0.7053, and Sampson error 3.0697$\rightarrow$3.0601. It is worse on the two remaining metrics: PSNR is 0.06 dB lower (22.3884 vs.\ 22.4476) and SSIM 0.0031 lower (0.7249 vs.\ 0.7280).

Relative to Wan, TinyVAE is better on FID (11.6306 vs.\ 11.6415), FVD (115.9042 vs.\ 116.5837), LPIPS (0.2356 vs.\ 0.2411), and PSNR (22.3884 vs.\ 22.3026, $+$0.09 dB), and worse on FDD (72.0696 vs.\ 68.7932), SSIM (0.7249 vs.\ 0.7254, $-$0.0005), Det-F1 (0.7053 vs.\ 0.7116), and Sampson error (3.0601 vs.\ 2.9908). The largest relative difference between the two decoders is in FDD ($+4.8\%$ for TinyVAE); on the other seven metrics the relative difference is at most $2.3\%$ (Sampson error). Replacing the 555M decoder by the 19M decoder therefore changes every generation metric by less than 5\% in the full chain.

\begin{table}[t]
\centering
\caption{Fixed-latent reconstruction on 248 clips of 81 frames: PSNR, SSIM, LPIPS, and char\_sim/box are computed against the ground-truth clips. Decode speedup and model-memory savings are ratios relative to Wan.}\label{tab:5}
\scriptsize
\setlength{\tabcolsep}{3pt}
\renewcommand{\arraystretch}{1.08}
\resizebox{\textwidth}{!}{%
\begin{tabular}{l c c c c c c }
\toprule
Model & PSNR$\uparrow$ & SSIM$\uparrow$ & LPIPS$\downarrow$ & char\_sim/box$\uparrow$ & Speedup$\uparrow$ & Mem.\ savings$\uparrow$ \\
\midrule
Wan & \best{32.5432} & \best{0.9006} & 0.0556 & \best{0.4715} & 1.0$\times$ (ref.) & 1.0$\times$ (ref.) \\
TAEHV super & 29.7920 & 0.8572 & 0.1011 & 0.2442 & \best{59.8$\times$} & \best{26.8$\times$} \\
TinyVAE & 31.8750 & 0.8865 & \best{0.0520} & 0.3334 & \best{59.8$\times$} & \best{26.8$\times$} \\
\bottomrule
\end{tabular}%
}
\end{table}

\textbf{Reconstruction.} On fixed-latent reconstruction (\tabref{tab:5}), TinyVAE is better than TAEHV super on all four metrics: PSNR 29.7920$\rightarrow$31.8750 dB ($+$2.08 dB), SSIM 0.8572$\rightarrow$0.8865 ($+$0.0293), LPIPS 0.1011$\rightarrow$0.0520 ($-$0.0491), and char\_sim/box 0.2442$\rightarrow$0.3334 ($+$0.0892). Measured against the TAEHV super--Wan difference on each metric, TinyVAE recovers 76\% of it in PSNR (2.08 of 2.75 dB), 68\% in SSIM (0.0293 of 0.0434), and 39\% in char\_sim/box (0.0892 of 0.2273), and exceeds it in LPIPS (0.0491 against a difference of 0.0455). The objective \eqrefn{eq:tinyvae} contains no text-specific term.

Relative to Wan, TinyVAE is 0.67 dB lower in PSNR (31.8750 vs.\ 32.5432), 0.0141 lower in SSIM (0.8865 vs.\ 0.9006), and 0.0036 lower, hence better, in LPIPS (0.0520 vs.\ 0.0556). The largest remaining difference is in char\_sim/box, 0.3334 for TinyVAE against 0.4715 for Wan ($-$0.1381); on this metric the 19M decoder remains below the 555M decoder.

\textbf{Efficiency.} Under the timing setup of \figref{fig:2}, TinyVAE decodes 59.8$\times$ faster than Wan with 26.8$\times$ lower model memory.

\subsection{Qualitative results}\label{sec:7.3}

Aggregate metrics do not reveal condition-specific failure modes or the mechanism of controllability. We therefore complement them with targeted qualitative examples: long-tail and multi-view generation, trajectory and layout interventions, cross-trajectory memory, one-step autoregression, 30 s (and longer) rollouts, and decoder comparisons. These figures are representative, not a statistical evaluation.

\subsubsection{Bidirectional generation}\label{sec:7.3.1}

\begin{figure}[tbp]
\centering
\includegraphics[width=0.835\textwidth,height=0.539\textheight,keepaspectratio]{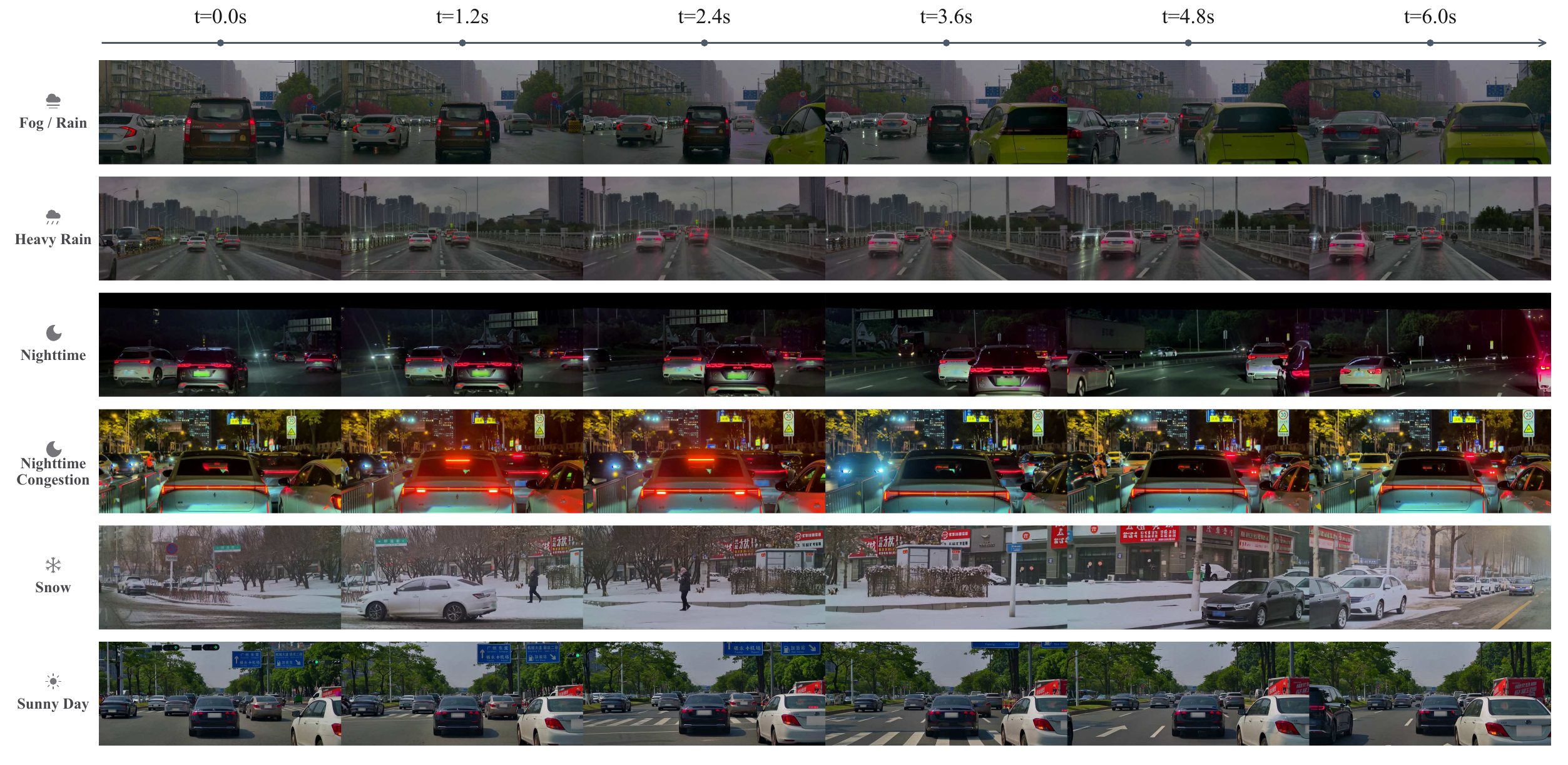}
\vspace{-1mm}
\caption{Front-narrow bidirectional generations across weather and lighting ($t=0$--$6.0$ s, $1.2$ s intervals).}\label{fig:19}
\end{figure}

\begin{figure}[tbp]
\centering
\includegraphics[width=0.800\textwidth,height=0.609\textheight,keepaspectratio]{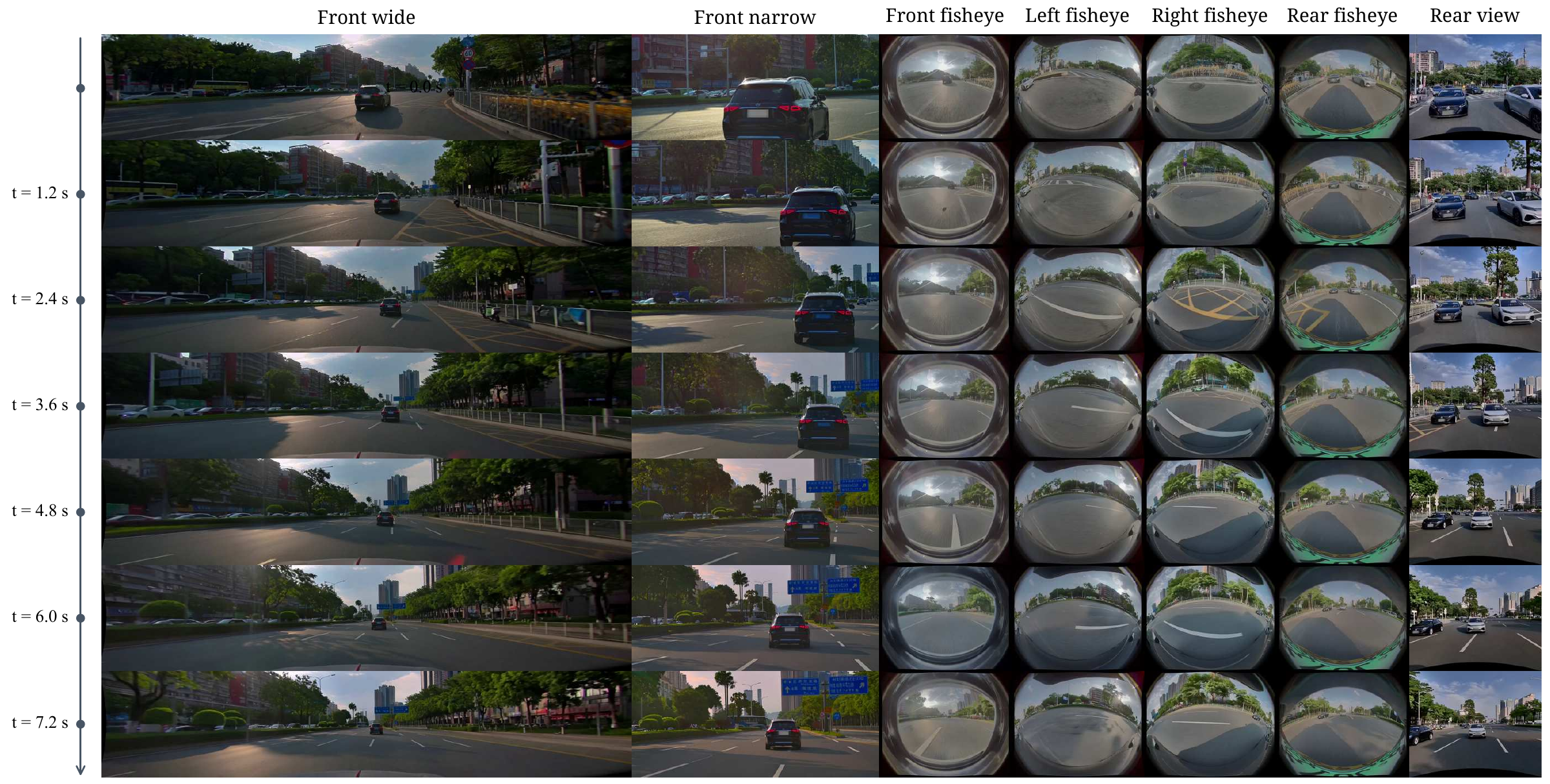}
\vspace{-1mm}
\caption{Synchronized seven-view bidirectional generations at seven timestamps ($t=0$--$7.2$ s, $1.2$ s intervals).}\label{fig:20}
\end{figure}

\figref{fig:19} shows front-narrow generations under rain and fog, heavy rain, night, congested night traffic, snow, and clear daytime, sampled every 1.2 s from 0 to 6.0 s. Wet-road reflections, precipitation occlusion, and headlight glare evolve smoothly, while vehicles and lane structure remain visually stable. \figref{fig:20} shows synchronized seven-view outputs at seven timestamps ($t=0$--$7.2$ s, $1.2$ s intervals); shared vehicles and road geometry stay consistent across fisheye and pinhole views. The examples sit beside the aggregate long-tail and geometry metrics; they do not replace them.

\subsubsection{Conditional controllability: trajectory and layout}\label{sec:7.3.2}

\begin{figure}[tbp]
\centering
\includegraphics[width=0.90\textwidth,height=0.58\textheight,keepaspectratio]{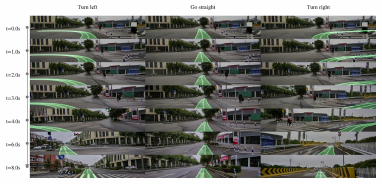}
\vspace{-1mm}
\caption{Trajectory intervention from a shared intersection (left, straight, right).}\label{fig:21}
\end{figure}

\begin{figure}[tbp]
\centering
\includegraphics[width=0.90\textwidth,height=0.58\textheight,keepaspectratio]{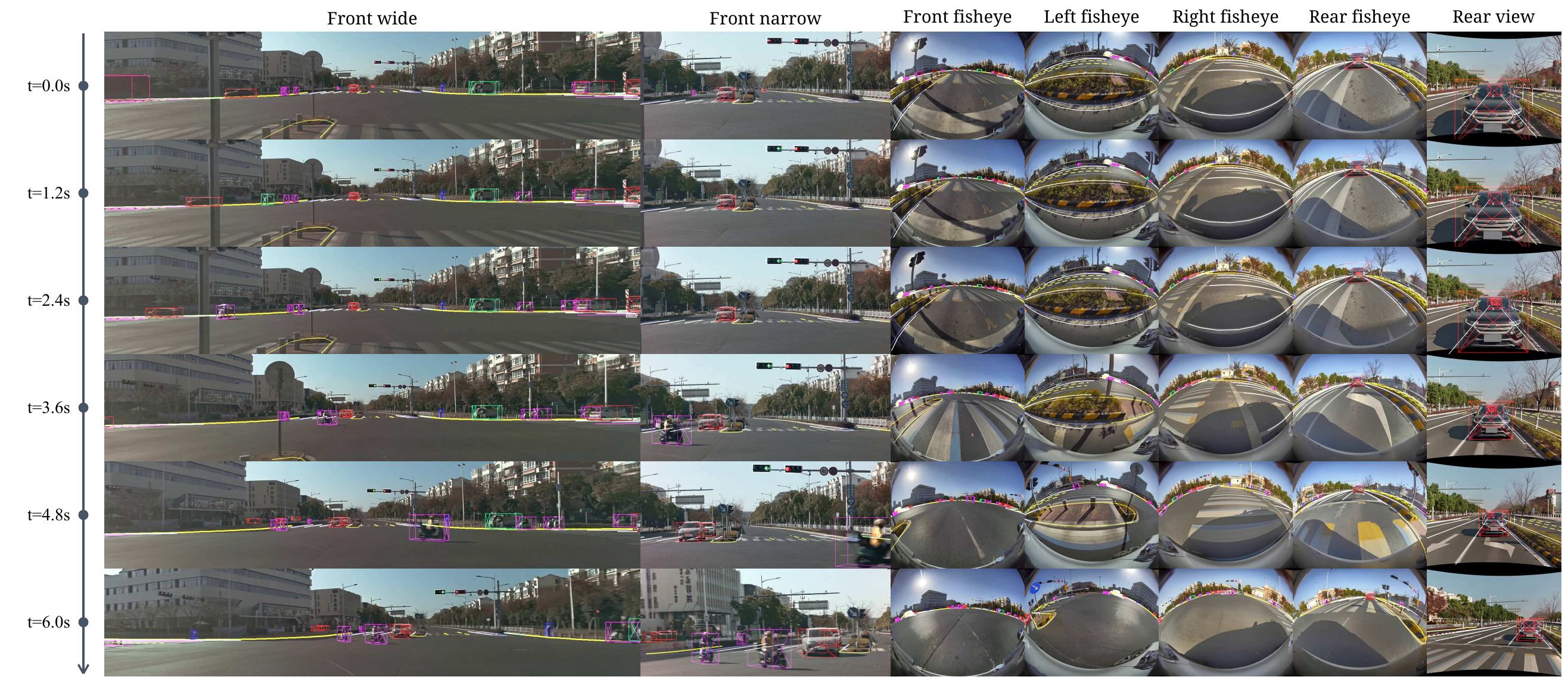}
\vspace{-1mm}
\caption{Seven-view layout following for vehicles and lane markings over time.}\label{fig:22}
\end{figure}

\begin{figure}[tbp]
\centering
\includegraphics[width=0.90\textwidth,height=0.58\textheight,keepaspectratio]{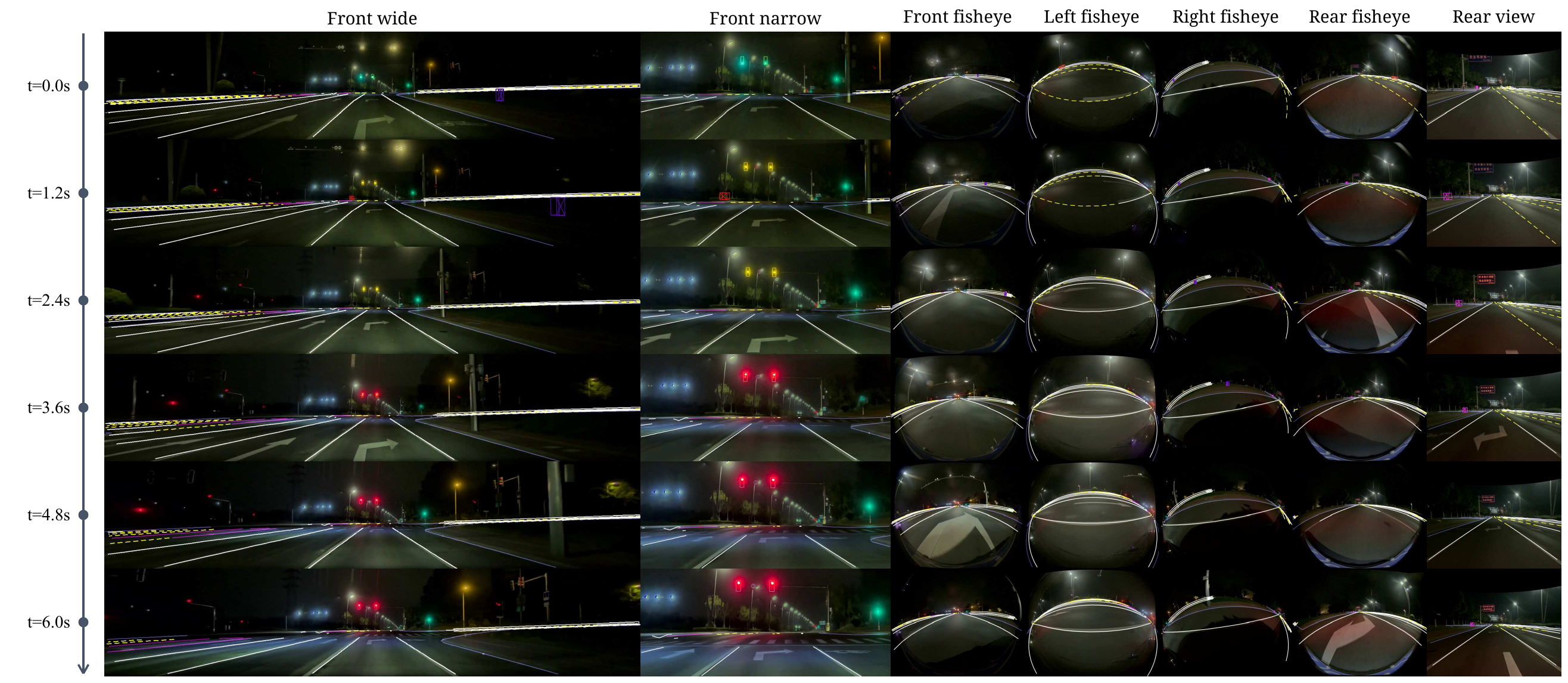}
\vspace{-1mm}
\caption{Traffic-signal intervention: green, yellow, and red phases with the ego vehicle stopping on red.}\label{fig:22b}
\end{figure}

\figref{fig:21} fixes the history and scene conditions while changing only the ego-motion command among left turn, straight, and right turn. The resulting road geometry and viewpoint evolution follow the commanded branch, providing a qualitative intervention test of trajectory control. \figref{fig:22} shows seven-view layout following over time: vehicles of different sizes and attributes and lane markings remain aligned with their wireframes. \figref{fig:22b} drives a single traffic light through green, yellow, and red within one rollout; the generated signal changes phase accordingly and the ego vehicle decelerates to a stop on red. These figures show controllability on selected examples, not exhaustive constraint satisfaction.

\subsubsection{Cross-trajectory memory}\label{sec:7.3.3}

\begin{figure}[tbp]
\centering
\includegraphics[width=0.835\textwidth,height=0.539\textheight,keepaspectratio]{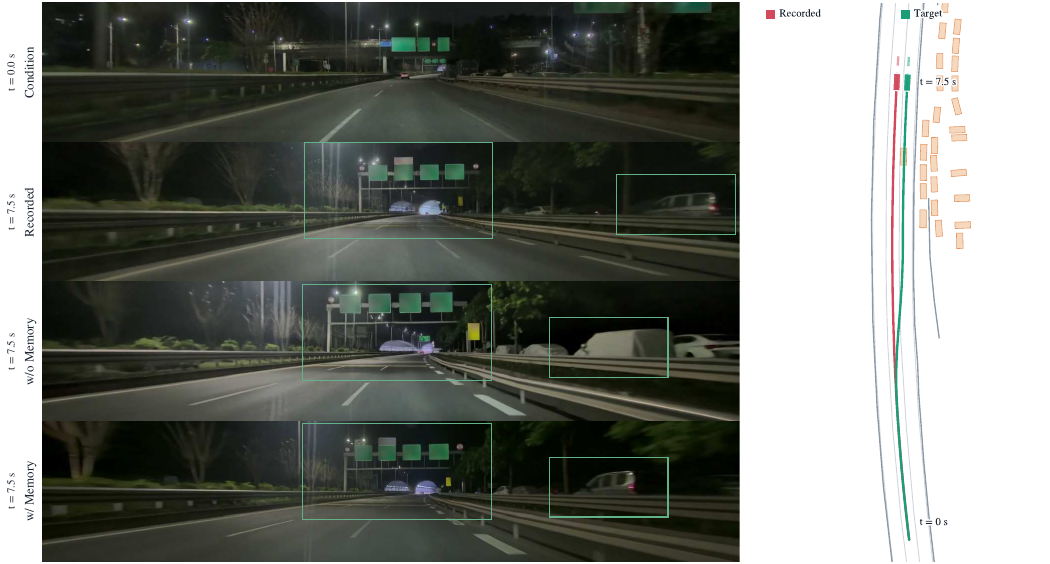}
\vspace{-1mm}
\caption{Cross-trajectory revisit with and without memory.}\label{fig:23}
\end{figure}

\figref{fig:23} evaluates implicit memory on a revisited road segment. Each group shows recorded-log memory, the target view from another trajectory, and generations without and with memory. Given conditioning frames and recorded-log memory from a different traversal, the model preserves place-specific details such as sign position and color and parked vehicles more faithfully than an unconstrained redraw. This is a qualitative revisit check; there is no standalone quantitative memory benchmark in this report.

\subsubsection{One-step autoregressive inference}\label{sec:7.3.4}

\begin{figure}[tbp]
\centering
\includegraphics[width=0.835\textwidth,height=0.539\textheight,keepaspectratio]{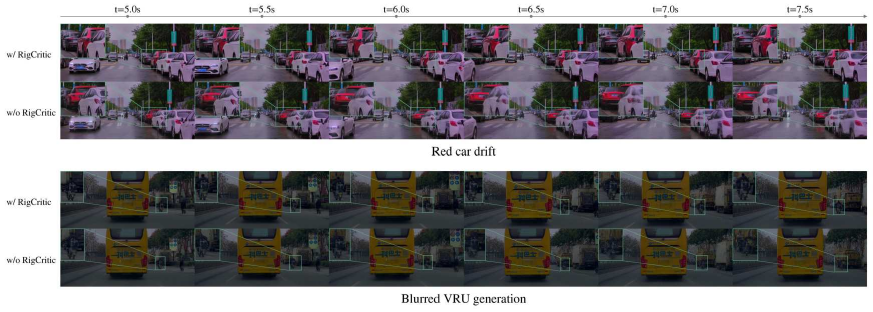}
\vspace{-1mm}
\caption{One-step autoregressive generation before and after RigCritic.}\label{fig:24}
\end{figure}

\figref{fig:24} compares one-step autoregressive outputs before and after RigCritic refinement. The refined model shows less visible drift and preserves road structure and multi-view appearance more consistently in the displayed sequence. This is a model-ablation visualization, not a deployment benchmark; one-step latency is in \secref{sec:7.2.3} and \figref{fig:2}.

\subsubsection{30 s long-horizon inference}\label{sec:7.3.5}

\begin{figure}[tbp]
\centering
\includegraphics[width=0.800\textwidth,height=0.609\textheight,keepaspectratio]{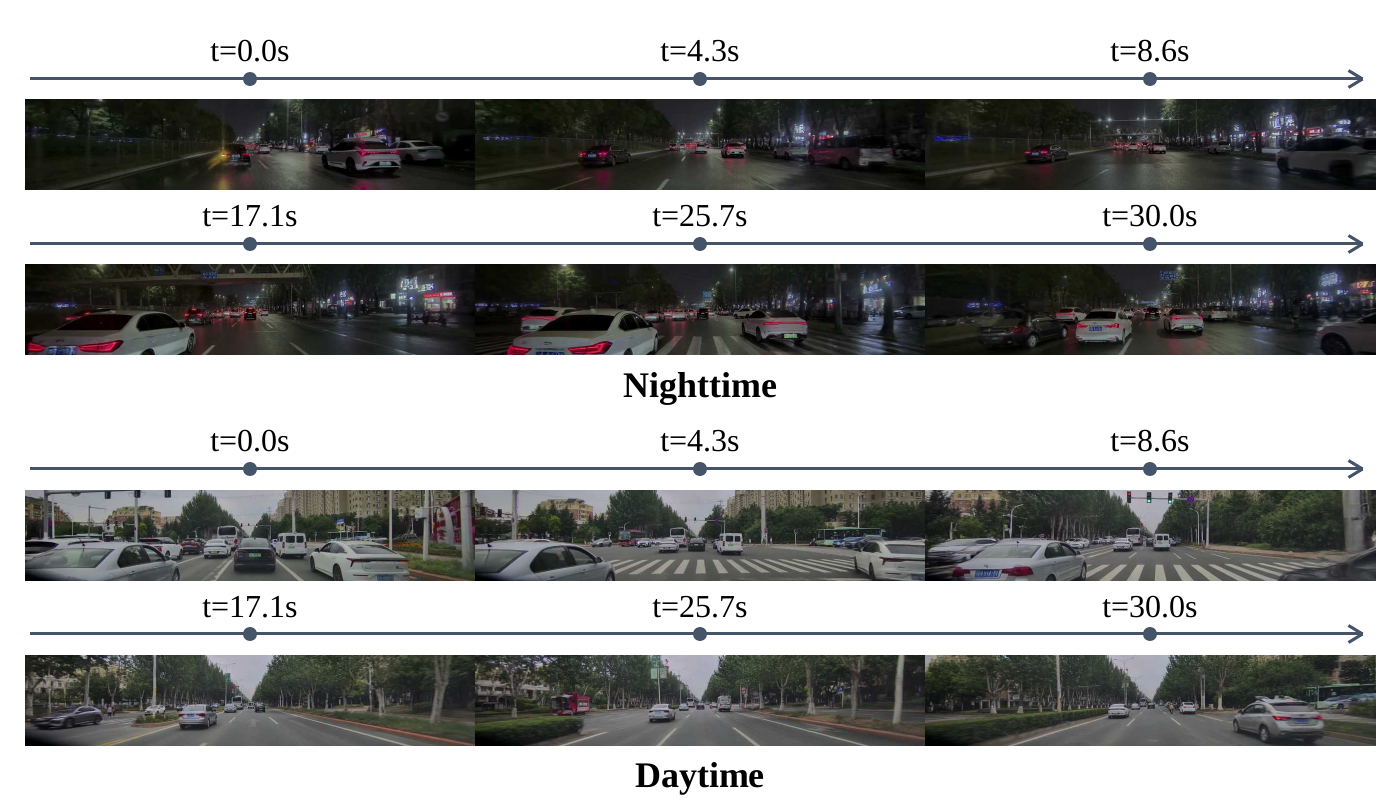}
\vspace{-1mm}
\caption{Two 30 s autoregressive rollouts, sampled non-uniformly at $t=0.0$, $4.3$, $8.6$, $17.1$, $25.7$, and $30.0$ s.}\label{fig:25}
\end{figure}

\figref{fig:25} shows two 30 s autoregressive rollouts (76 latent timesteps) under the bounded KV-cache. Scene layout, lane topology, and illumination remain visually stable in these examples. The same streaming loop continues at minute scale; the figure reports 30 s as a compact qualitative horizon, and a larger long-horizon stability study is left open.

\subsubsection{Decode fidelity of TinyVAE}\label{sec:7.3.6}

\begin{figure}[tbp]
\centering
\includegraphics[width=0.835\textwidth,height=0.539\textheight,keepaspectratio]{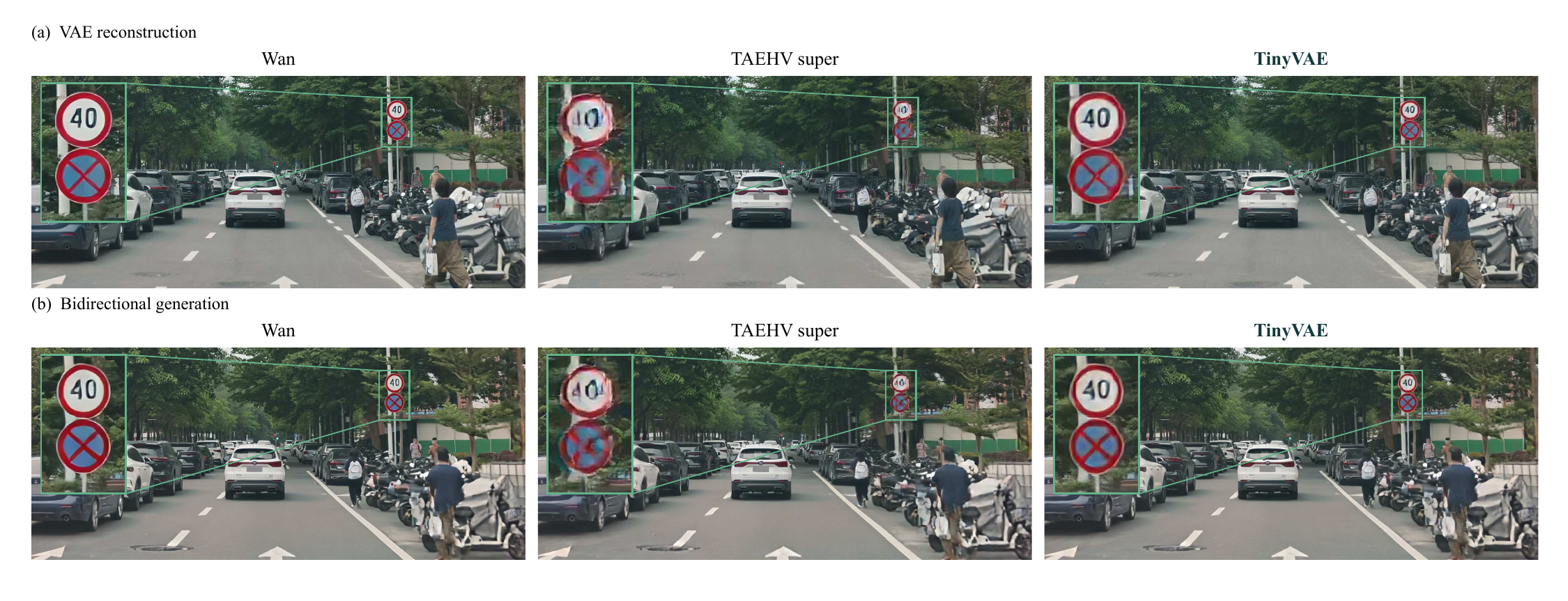}
\vspace{-1mm}
\caption{Same-latent reconstruction (a) and bidirectional generation (b): Wan, TAEHV super, and TinyVAE.}\label{fig:26}
\end{figure}

\figref{fig:26} compares Wan, TAEHV super, and TinyVAE for fixed-latent reconstruction and same-latent bidirectional generation. TinyVAE approaches Wan more closely than TAEHV super, particularly for text and small structures, while using less than 4\% of the parameters and delivering the \best{59.8$\times$} decoding speedup measured under \figref{fig:2}'s timing protocol. The gaps in \secref{sec:7.2.4} still stand: visual similarity here is not numerical equivalence.

\FloatBarrier
\section{Conclusion}\label{sec:8}

A world model used as the policy's sensor stream has to match the production interface, geometry, timing, and compute budget closely enough to stand in for part of on-road testing. ZYT-World is built for a mixed production rig and puts seven-view joint modeling, pixel-aligned control, per-latent causal generation, cross-trajectory memory, and real-time decoding on one chain. Four ultra-wide fisheye cameras and three pinhole cameras keep native projections; Pl\"ucker rays, ego-motion AdaLN, and layout inject camera, ego-vehicle, and scene conditions. Three-stage autoregressive distillation compresses a multi-step bidirectional teacher into a one-step streaming student; RigCritic refines the full rig, and a bounded KV-cache carries generation forward. Cross-trajectory paired captures and a zero-initialized residual path give plug-in memory; TinyVAE supplies both real-time decoding and differentiable perceptual supervision. Heterogeneous scheduling, multi-dimensional parallelism, W8A8, and our inference engine make seven-view high-resolution training and few-step deployment feasible. The experiments cover generation quality, reconstruction fidelity, and downstream usability.

\section*{Contributors}

Turning a world model into a production-grade closed-loop simulator required joint progress on data, algorithms, training, infrastructure, and evaluation. We thank every member of the team for the sustained effort that made ZYT-World possible.

\vspace{0.8em}
\noindent\textbf{Advisers:} Kaixuan Wang, Zichao Guo, Xiaozhi Chen\\[0.7em]
\textbf{Project Lead:} Wei Bi\\[0.7em]
\textbf{Contributors:} Boni Hu\textsuperscript{*}, Xiong Wei\textsuperscript{*}, Haoming Huang\textsuperscript{*}, Yong Huang\textsuperscript{*}, Chenbo Wang\textsuperscript{*}, Yi Yang, Jiancheng Wang\textsuperscript{\ddag}, Ruicheng Zhu, Zhimin Yang, Guanglai Liu, Qiaowan Jin\textsuperscript{\ddag}, Dongzhuo Wang\textsuperscript{\ddag}, Haiwei Kuang, Jiajun Fan, Yue Wu\textsuperscript{\ddag}, Jiaxin Wei\textsuperscript{\ddag}, Hao Sun, Feihong Yan\textsuperscript{\ddag}, Yuyao Zhou\\[0.9em]
{\small
\noindent\textsuperscript{*}~Core contribution.\\
\textsuperscript{\ddag}~Research Intern at ZYT.}

\nocite{yang2026dreamerad,yan2026causaldrive}
\bibliography{biblio}
\bibliographystyle{assets/plainnat}
\end{document}